%% file: iclr2027_conference.tex
\documentclass{article} % For LaTeX2e
\usepackage{iclr2027_conference,times}

\input{math_commands.tex}

\usepackage{graphicx}
\graphicspath{{figs/}{../figs/}}
\usepackage{booktabs}
\usepackage{amsmath}
\usepackage{amssymb}
\usepackage{xcolor}
\usepackage{colortbl}
\usepackage{array}
\usepackage{tabularx}
\newcolumntype{Y}{>{\centering\arraybackslash}X}
\definecolor{pdmscol}{gray}{0.92}
\colorlet{defrow}{gray!20}
\definecolor{catrow}{HTML}{EEE5F4}
\newcolumntype{P}{c}
\newcolumntype{Q}{c}
\usepackage{placeins}
\usepackage{subcaption}
\usepackage{hyperref}
\usepackage{url}
\title{PlanWAM: Planning-Shaped Future Representations for End-to-End Autonomous Driving}

\author{%
\textbf{Jinchang~Xu\textsuperscript{1,\dag},
Hongda~Yu\textsuperscript{2,\dag},
Fengwei~Dong\textsuperscript{1},
Wenhui~Huang\textsuperscript{3},
Xi~Wei\textsuperscript{2,\ddag},
Yongzhi~Liu\textsuperscript{1},\\
Sunan~Zhang\textsuperscript{1},
Jirao~Wang\textsuperscript{4},
Chen~Lv\textsuperscript{3},
Bingbing~Li\textsuperscript{1},
Guodong~Yin\textsuperscript{1},
Weichao~Zhuang\textsuperscript{1,*}}\\[0.45em]
\mdseries\normalsize
\textsuperscript{1}\,Southeast University,
\textsuperscript{2}\,NIO,
\textsuperscript{3}\,Nanyang Technological University,
\textsuperscript{4}\,Tongji University\\[0.35em]
Project Page: \url{https://jinchan9.github.io/PlanWAM_project_page}
}

\iclrfinalcopy
\makeatletter
\def\@maketitle{\vbox{\hsize\textwidth
{\LARGE\sc \@title\par}
\vskip 0.16in
\ificlrfinal
  \lhead{Published as a conference paper at ICLR 2027}%
  \centering\@author
\else
  \lhead{Under review as a conference paper at ICLR 2027}%
  \centering\textbf{Anonymous authors}\\Paper under double-blind review
\fi
\par\vskip 0.22in}}
\makeatother
\begin{document}

\maketitle
\lhead{Preprint}
\makeatletter
\begingroup
\renewcommand{\thefootnote}{}%
\renewcommand{\@makefntext}[1]{\noindent #1}%
\footnotetext{$^{\dagger}$These authors contributed equally to this work. Work done during an internship at NIO.}%
\footnotetext{$^{\ddagger}$Project leader.}%
\footnotetext{$^{*}$Corresponding author: \texttt{wezhuang@seu.edu.cn}}%
\endgroup
\makeatother

\begin{abstract}
  %端到端自动驾驶中的世界模型通过预测未来场景演化，为轨迹规划提供前瞻信息。现有方法主要研究如何预测未来以及如何利用未来，而较少追问：何种未来表征才真正最有利于规划。为此，我们提出 PlanWAM，一种 Planning-Shaped World-Action Model。其核心思想是让规划任务塑造未来状态表征，使其保留对规划最有用的信息。随后由 latent world model 从历史观测中预测这一 planning-oriented future latent representation，并将其用于规划，从而实现具有前瞻性的规划。具体而言，我们首先使用 Temporal Register Pyramid，以 recency-aware 的方式压缩多帧历史信息，学习面向未来推理与规划的紧凑历史表征。随后引入可观测真实未来帧的 privileged future posterior，并用轨迹规划目标塑造其 future latent representation，得到 planning-oriented future latent representation。Hindsight-to-Foresight Distillation 训练一个仅依赖历史的  prior，以预测这一 future latent representation。预测得到的 future latent representation 作为 planning context，引导轨迹生成与选择。PlanWAM 在 NAVSIM-v1/v2 navtest 上取得 93.8 PDMS / 90.9 EPDMS，并在闭环 HUGSIM 的 zero-shot 设置下达到 38.7 HD-Score，在开环与闭环评测中均展现出领先的规划表现。大量实验进一步表明，planning-shaped future representations 为 world-action model 提供了一种有效且可部署的 foresight。
  World models in end-to-end autonomous driving predict future scene evolution to provide foresight for trajectory planning. Existing methods mainly study how to predict the future and how to use it, but less often ask which future representation is actually most useful for planning. To this end, we propose PlanWAM, a Planning-Shaped World Action Model. The key idea is to let the planning task shape the future-state representation, so that it retains the information most useful for planning. A latent world model then predicts this planning-shaped future latent representation from historical observations and uses it for planning, enabling foresighted planning. Specifically, we first use a Temporal Register Pyramid to compress multi-frame historical information in a recency-aware manner, learning a compact history representation oriented toward future reasoning and planning. We then introduce a privileged future posterior branch that observes ground-truth future frames, and shape its future latent representation with trajectory-planning objectives to obtain a planning-shaped future latent representation. Hindsight-to-Foresight Distillation trains a prior branch that depends only on history to predict this future latent representation. The predicted future latent representation serves as planning context and guides trajectory generation and selection. PlanWAM achieves 93.8 PDMS / 90.9 EPDMS on NAVSIM-v1/v2 navtest and reaches 38.7 HD-Score on closed-loop HUGSIM in a zero-shot setting, demonstrating leading planning performance across both open-loop and closed-loop evaluations. Extensive experiments further demonstrate that planning-shaped future representations provide an effective and deployable form of foresight for world-action models.
\end{abstract}

\section{Introduction}
%端到端自动驾驶通过统一模型直接从原始传感器观测规划未来轨迹，使场景表征能够围绕最终规划目标进行学习。然而，可靠的驾驶规划不仅依赖当前观测，还需要利用历史信息辨识交通参与者的运动趋势与交互关系，并推断场景未来的可能演化。由此，世界模型逐渐成为端到端自动驾驶的重要研究方向：通过建模场景随时间及车辆行为的演化，为规划器提供未来信息，使轨迹规划从响应当前场景走向前瞻性规划。
End-to-end autonomous driving uses a unified model to plan future trajectories directly from raw sensor observations, so that scene representations can be learned for the final planning objective~\cite{codevilla2018end,hu2023planning,jiang2023vad,liao2025diffusiondrive}. Reliable driving, however, depends on more than the current observation. A planner must also use history to identify motion trends and interactions among traffic participants, and to infer possible future scene evolution. World models have therefore become an important research direction for end-to-end autonomous driving~\cite{wang2024drivedreamer,zheng2024occworld,zhou2025hermes}. By modeling how the scene evolves over time and with vehicle behavior, they supply future information to the planner, shifting trajectory planning from reacting to the current scene toward foresighted planning.

\begin{figure}[h]
  \centering
  \includegraphics[width=1.0\linewidth]{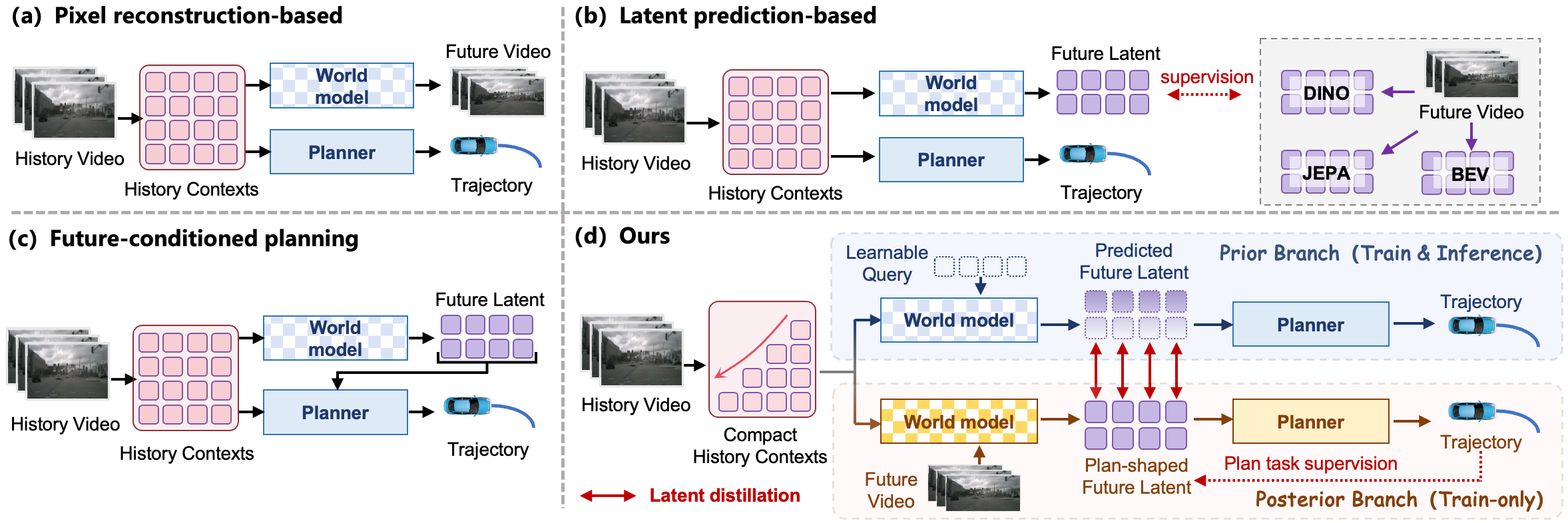}
  %图1：驾驶 world-action models (WAM) 的四种未来建模范式。(a) Reconstruction；(b) Latent prediction；(c) Future-conditioned planning；(d) PlanWAM (ours)。
  \caption{\textbf{Four future-modeling paradigms in driving World Action Models (WAMs).} (a)~Pixel reconstruction. (b)~Latent prediction. (c)~Future-conditioned planning. (d)~PlanWAM (ours).}
  \label{fig:intro}
\end{figure}

%现有 driving world models 在未来信息的表示与利用方式上，逐渐从显式未来重建发展到 latent prediction，并进一步走向 future-conditioned planning，如 Fig. 1 所示。早期方法通过生成未来图像、视频或稠密场景状态学习环境动态，如 Epona、PWM 等，但往往需要建模大量与规划弱相关的视觉细节，并伴随较高的生成成本。随后，LAW、Latent-WAM 等转向 latent world modeling，在紧凑特征空间中预测未来状态，但其未来表征主要作为预测目标或辅助监督，对规划的作用仍较间接。近期的 DriveFuture、WorldDrive 等进一步将 predicted future latent 显式引入 trajectory planner，使未来信息直接参与规划。尽管如此，现有方法大多关注如何预测未来以及如何利用未来，而较少研究一个更基础的问题：何种 future representation 才最有利于规划？现有未来表征通常由通用特征预测、视觉重建或预设状态目标所定义，因此可能与下游 planning objective 存在表征错配：一个能够重建场景的未来表征，并不一定保留了影响轨迹决策的关键信息。
Existing driving world models have gradually evolved in how they represent and use future information, moving from pixel reconstruction to latent prediction and then to future-conditioned planning (Fig.~\ref{fig:intro}). Early methods, such as Epona~\cite{zhang2025epona} and PWM~\cite{zhao2026forecasting}, learn environment dynamics by generating future images, videos, or dense scene states, but they often have to model substantial visual detail that is only weakly related to planning, at a high generation cost. Later methods, such as LAW~\cite{li2025law} and Latent-WAM~\cite{wang2026latentwam}, turn to latent world modeling and predict future states in a compact feature space. Their future representations mainly serve as prediction targets or auxiliary supervision, so the effect on planning remains indirect. Recent methods, such as DriveFuture~\cite{hong2026drivefuture} and WorldDrive~\cite{gui2026worlddrive}, further introduce the predicted future latent explicitly into the trajectory planner, so that future information can participate in planning directly. Even so, most existing methods focus on how to predict the future and how to use it, and less often study a more basic question: \emph{which future representation is most useful for planning?} Existing future representations are usually defined through pixel-level reconstruction or image latent prediction, such as DINO, JEPA, or BEV feature, and may therefore be mismatched with the downstream planning objective: a future representation that can reconstruct the scene does not necessarily retain the key information that affects trajectory decisions.

%我们将这一问题重新表述为一个面向规划的 predictive representation learning 问题：世界模型不仅需要预测未来，还需要学习哪些未来信息值得被预测。对于自动驾驶而言，有效的表征无需完整保留历史观测或重建未来场景，而应聚焦于与轨迹规划相关的信息。历史表征应从冗余的多帧多视角观测中提炼支持未来推理与规划的动态证据；未来表征则应由 trajectory planning objectives 直接塑造，使其优先编码真正影响轨迹决策的信息。进一步地，这一 planning-oriented future representation 应能够由历史观测预测，并作为 foresight 参与轨迹规划。基于这一视角，我们将历史建模、未来预测与轨迹规划统一为一条 planning-oriented predictive representation learning 链路：由规划定义有用的未来，由未来约束历史信息的保留，并利用预测得到的未来表征改善规划。
We reformulate future modeling in end-to-end driving as a planning-oriented predictive representation learning problem: a world model must not only predict the future, but also learn which future information is worth predicting. For autonomous driving, an effective representation need not fully retain historical observations or reconstruct the future scene, but should focus on information relevant to trajectory planning. The history representation should capture dynamic cues from multi-frame, multi-view observations that are essential for future reasoning and planning, while the future representation should be shaped directly by trajectory planning objectives to prioritize decision-relevant future information. Furthermore, a future representation intended for planning should be predictable from historical observations and serve as foresight for trajectory planning. From this perspective, we unify historical modeling, future prediction, and trajectory planning into a planning-shaped predictive representation learning framework, where planning defines what future information matters, future prediction guides which historical information is retained, and the resulting future representation in turn improves planning.

%基于这一观点，我们提出 PlanWAM，一种 Planning-Shaped World-Action Model，如 Fig. 1(d) 所示。首先，我们以 recency-aware 的方式压缩多帧多视角历史观测，并在未来预测与规划目标的联合监督下学习紧凑的 history representation，使其优先保留支持 future reasoning 的关键动态信息。随后，我们构建一个仅在训练阶段使用的 privileged future posterior，利用真实未来观测构造 future latent representation，并通过 trajectory planning objectives 塑造其表示空间，得到 planning-oriented future latent representation。由于真实未来在部署阶段不可获得，我们进一步提出 Hindsight-to-Foresight Distillation，训练一个仅依赖历史信息的  prior 来预测该 future representation，从而将训练阶段的 privileged hindsight 转化为部署阶段可获得的 foresight。预测得到的 future representation 作为前瞻性 planning context，与 history representation 融合以指导轨迹生成与选择，实现 Foresight-Conditioned Planning。我们的主要贡献如下：
Building on this perspective, we propose PlanWAM, a Planning-Shaped World Action Model, as illustrated in Fig.~\ref{fig:intro}(d). First, we compress multi-frame, multi-view historical observations in a recency-aware manner and learn a compact history representation under joint future-prediction and planning supervision, preserving key dynamic cues for future reasoning. We then introduce a training-only privileged future posterior branch that leverages ground-truth future observations to construct a future latent representation, whose representation space is shaped by trajectory planning objectives to obtain a planning-shaped future latent representation. Since future observations are unavailable at deployment, Hindsight-to-Foresight Distillation trains a prior branch conditioned only on history to predict this representation, thereby converting privileged hindsight during training into deployable foresight at inference time. The predicted future representation is fused with the history representation as a foresight planning context to guide trajectory generation and selection, enabling Foresight-Conditioned Planning. Our main contributions are as follows:
\begin{itemize}
    %（1）我们将端到端驾驶中的未来建模重新表述为一个面向规划的 predictive representation learning 问题：future representation 不再由预设的重建或通用预测目标定义，而是由 trajectory planning objectives 直接塑造，使其优先编码影响轨迹决策的未来信息。
    \item We reformulate future modeling in end-to-end driving as a planning-oriented predictive representation learning problem, where the future representation is shaped directly by trajectory planning objectives rather than predefined reconstruction or generic prediction objectives, so that it preferentially encodes decision-relevant future information.
    %（2）我们提出 PlanWAM：提炼支持 future reasoning 的紧凑历史表征，利用真实未来观测与规划目标学习 planning-shaped future latents，并通过 Hindsight-to-Foresight Distillation 从历史预测该表征，作为 foresight 用于轨迹生成与选择。
    \item We propose PlanWAM, which extracts a compact history representation for future reasoning, learns planning-shaped future latents from ground-truth future observations and planning objectives, and uses Hindsight-to-Foresight Distillation to predict this representation from history as foresight for trajectory generation and selection.
    %（3）PlanWAM 在 NAVSIM-v1/v2 navtest 上达到 93.8 PDMS / 90.9 EPDMS，并在 HUGSIM zero-shot 闭环评测中达到 38.7 HD-Score。进一步分析表明，planning-shaped future representations 能够改善轨迹生成与选择，并减少碰撞和低 TTC 等安全关键失败。
    \item PlanWAM achieves 93.8 PDMS / 90.9 EPDMS on NAVSIM-v1/v2 navtest and 38.7 HD-Score in zero-shot closed-loop evaluation on HUGSIM. Further analyses show that planning-shaped future representations improve trajectory generation and selection while reducing safety-critical failures, including collisions and low-TTC events.
\end{itemize}

\section{Related Work}
\label{sec:related}

\subsection{End-to-End Autonomous Driving}
%端到端自动驾驶通过统一模型直接从原始传感器观测规划未来轨迹，以减少传统模块化系统中的信息损失与级联误差。早期方法以结构化场景表示与多任务联合学习为主，将感知、预测与规划纳入同一模型，如 UniAD、VAD、SparseDrive 和 TransFuser 等。随后，研究转向多模态与生成式轨迹规划，以刻画复杂交通中多样且不确定的驾驶行为，如 iPad、DrivoR、DiffusionDrive 和 GoalFlow 等。近期方法进一步引入视觉语言模型与视觉语言动作模型，如 ReCogDrive、AutoVLA、Orion 和 Qwen-Drive-1.0 等，以增强语义理解与推理。尽管这些方法不断提升场景表征与轨迹规划能力，其决策仍主要建立在当前或历史观测之上，对未来场景演化的利用仍相对有限。相比之下，PlanWAM 用世界模型预测对规划有用的未来表征，并以此引导轨迹生成与评估。
End-to-end autonomous driving uses a unified model to plan future trajectories directly from raw sensor observations, thereby reducing information loss and cascading errors in traditional modular systems. Early methods, such as UniAD~\cite{hu2023planning}, VAD~\cite{jiang2023vad}, SparseDrive~\cite{sun2025sparsedrive}, and TransFuser~\cite{chitta2022transfuser}, used structured scene representations and multi-task learning to jointly model perception, prediction, and planning. Subsequent methods, such as iPad~\cite{guo2025ipad}, DrivoR~\cite{kirby2026drivor}, DiffusionDrive~\cite{liao2025diffusiondrive}, and GoalFlow~\cite{xing2025goalflow}, adopt multimodal and generative trajectory planning to capture diverse and uncertain driving behaviors. More recent methods, such as ReCogDrive~\cite{chen2026recogdrive}, AutoVLA~\cite{autovala2025}, Orion~\cite{fu2025orion}, and Qwen-Drive-1.0~\cite{zhou2026qwendrive}, introduce vision-language and vision-language-action models to strengthen semantic understanding and reasoning. Although these methods keep improving scene representation and trajectory planning, they still decide mainly from current or historical observations and make limited use of future scene evolution. In contrast, PlanWAM uses a world model to predict a future representation useful for planning, and guides trajectory generation and evaluation with that foresight.

\subsection{World Models for Autonomous Driving}
%世界模型通过建模场景随时间及车辆行为的演化，为端到端驾驶提供前瞻信息。现有 driving world models 一类侧重于未来真实世界的生成与模拟，如 DriveDreamer、Drive-WM、Vista、Epona 和 PWM 等，通过预测图像、视频或 BEV 等结构化场景状态来学习交通演化，并主要关注生成质量、时序一致性、动作可控性与长时预测。另一类工作更关注 world modeling 对驾驶任务本身的作用，如 LAW、Latent-WAM、World4Drive、DriveFuture 和 WorldDrive 等，通过预测未来特征或 latent states 学习动态场景表征，并进一步辅助或直接参与轨迹规划。上述研究表明，未来建模能够为端到端驾驶提供有效的动态先验，但现有工作主要关注如何更好地预测未来特征以及如何利用未来信息，而较少从 planning objective 本身出发研究 future representation 应编码哪些信息。相比之下，PlanWAM 研究何种未来表征才真正最有利于规划，直接用规划目标塑造 future latent representation，使其重点提取与规划决策相关的关键未来信息。
World models provide foresight for end-to-end driving by modeling how the scene evolves over time and under vehicle behavior. Existing driving world models fall into two lines. One line, such as DriveDreamer~\cite{wang2024drivedreamer}, Drive-WM~\cite{wang2024drivewm}, Vista~\cite{gao2024vista}, Epona~\cite{zhang2025epona}, and PWM~\cite{zhao2026forecasting}, focuses on generating and simulating the future real world. These methods learn traffic evolution by predicting images, videos, or structured scene states such as BEV, and mainly study generation quality, temporal consistency, action controllability, and long-horizon prediction. The other line, such as LAW~\cite{li2025law}, Latent-WAM~\cite{wang2026latentwam}, World4Drive~\cite{zheng2025world4drive}, DriveFuture~\cite{hong2026drivefuture}, and WorldDrive~\cite{gui2026worlddrive}, focuses on the role of world modeling in the driving task itself, learning dynamic scene representations by predicting future features or latent states and further assisting or directly participating in trajectory planning. These studies show that future modeling can provide a useful dynamic prior for end-to-end driving. Existing work, however, mainly studies how to predict future features more effectively and how future information is used, and less often asks, from the planning objective itself, what the future representation should encode. In contrast, PlanWAM studies which future representation is most useful for planning and shapes the future latent representation directly with planning objectives, so that the latent preferentially extracts the key future information relevant to planning decisions.

\section{Method}
\label{sec:method}

\subsection{Problem Formulation and Framework Overview}
\label{sec:overview}
%设 $v_{t-n:t}$ 表示历史多视角相机观测，$c_{t-n:t}$ 表示对应的自车状态与导航信息，$\tau_{1:T}$ 表示待规划的未来自车轨迹。常规端到端规划器直接建模 $p(\tau_{1:T}\mid v_{t-n:t},c_{t-n:t})$。
Let $v_{t-n:t}$ denote the historical multi-view camera observations, $c_{t-n:t}$ the corresponding ego states and navigation information, and $\tau_{1:T}$ the future ego trajectory to be planned. A conventional end-to-end planner directly models:
\begin{equation*}
    p(\tau_{1:T} \mid v_{t-n:t}, c_{t-n:t}).
\end{equation*}

\begin{figure}[h]
   \centering
   \includegraphics[width=1.0\linewidth]{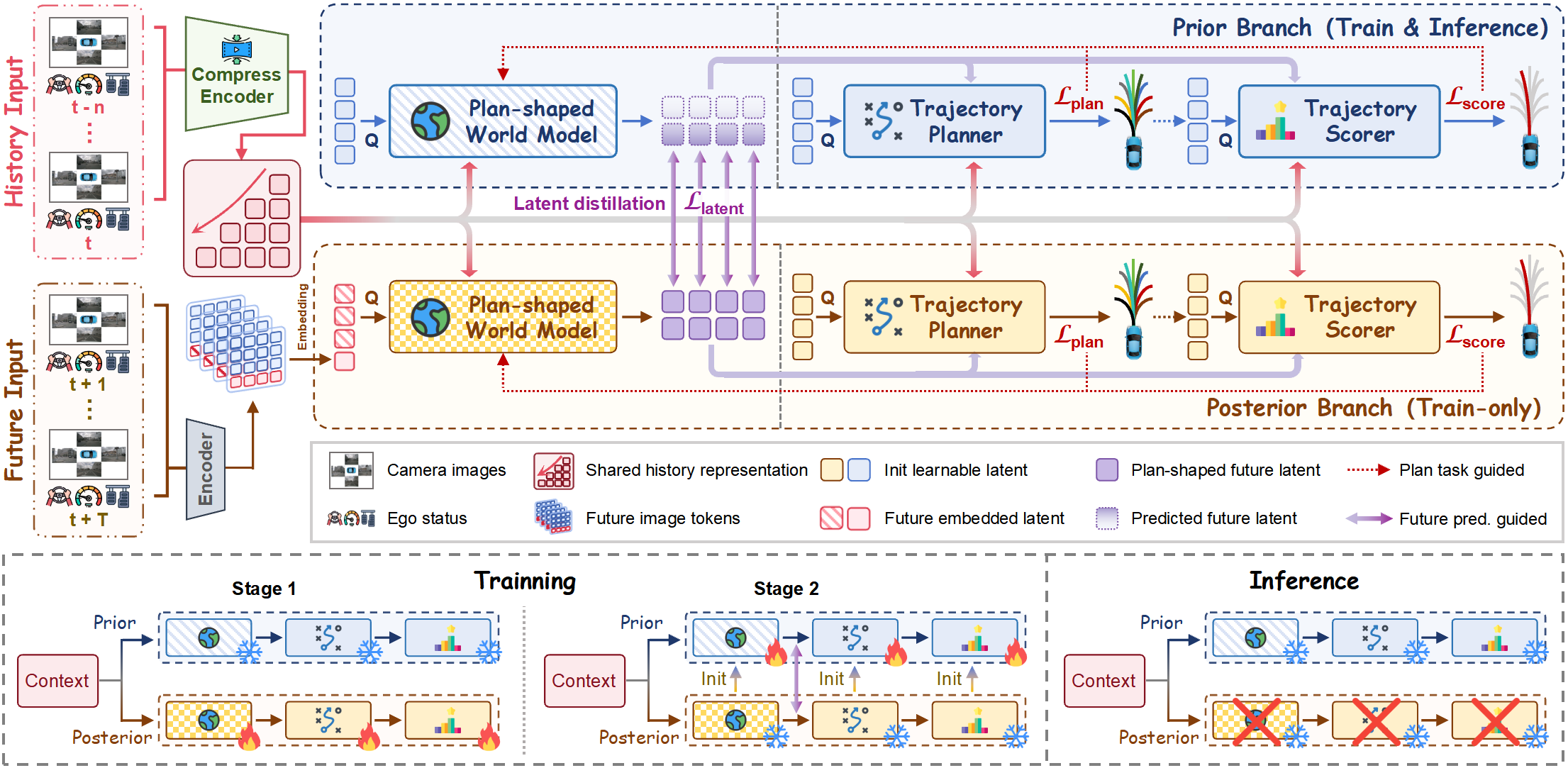}
   %图2：PlanWAM 总览。多视角历史观测被压缩为支持未来推理的紧凑表征。特权 posterior 分支利用真实未来观测与规划目标学习 planning-shaped future latents，并蒸馏到仅依赖历史的  prior。预测得到的 future latents 作为 foresight，用于轨迹生成与选择。posterior 仅用于训练，推理只依赖 prior 分支。
   \caption{
    \textbf{Overview of PlanWAM}. 
    Historical multi-view observations are compressed into a compact representation for future reasoning.
    The privileged posterior branch uses ground-truth future observations and planning objectives to learn planning-shaped future latents, which are distilled into a prior branch conditioned only on history.
    The predicted future latents then serve as foresight for trajectory generation and selection. The posterior is training-only; inference uses only the prior.
    }
   \label{fig:overview}
\end{figure}

%为显式利用未来场景演化，future-conditioned world-action models 在历史观测与轨迹规划之间引入未来状态表征 $z_t^{+}$，并将规划过程分解为：
To use future scene evolution explicitly, future-conditioned WAMs (Fig.~\ref{fig:intro}(c)) introduce a future-state representation $z_{t}^{+}$ between historical observations and trajectory planning, and factorize the process as:
\begin{equation}
    p(\tau_{1:T}, z_{t}^{+} \mid v_{t-n:t}, c_{t-n:t})
    =
    p(z_{t}^{+} \mid v_{t-n:t}, c_{t-n:t})
    \, p(\tau_{1:T} \mid v_{t-n:t}, c_{t-n:t}, z_{t}^{+}).
    \label{eq:future_conditioned_planning}
\end{equation}
%这一分解使未来信息能够直接参与规划，但也引出了一个关键问题：应该如何定义和构造 $z_t^{+}$，才能使其对轨迹规划更有益？现有方法通常通过通用特征预测、视觉重建或预设状态目标定义未来表征，但这样的未来表征并不一定最有利于规划。为此，我们提出 PlanWAM，如图\ref{fig:overview}所示。
This factorization lets future information participate in planning, but it also raises a key question: \emph{how should $z_{t}^{+}$ be defined and constructed so that it is more beneficial for trajectory planning?} Existing methods typically define the future representation by pixel reconstruction or image latent prediction, such as DINO, JEPA, or BEV feature, but such a representation is not necessarily the most useful for planning. To this end, we propose PlanWAM, as illustrated in Fig.~\ref{fig:overview}.

%首先，我们采用历史压缩模块 $E_{\eta}$，以 recency-aware 的方式压缩多帧多视角历史观测，得到紧凑历史表征。
We first use a history compression module $E_{\eta}$ to compress multi-frame, multi-view historical observations in a recency-aware manner, yielding a compact history representation:
\begin{equation}
    h_{t} = E_{\eta}(v_{t-n:t}, c_{t-n:t}).
    \label{eq:history_representation}
\end{equation}

%在训练阶段，我们进一步引入可访问真实未来观测的 privileged future posterior $P_{\phi}$，构造 future latent representation：
During training, we further introduce a privileged posterior branch $P_{\phi}$, hereafter referred to as the posterior, which accesses ground-truth future observations and constructs a future latent representation:
\begin{equation}
    z_{t}^{+}
    =
    P_{\phi}\bigl(h_{t},\, v_{t+1:t+T},\, c_{t+1:t+T}\bigr).
    \label{eq:posterior_future_latent}
\end{equation}

%不同于通过辅助 reconstruction 或通用 prediction objective 预先规定 $z_t^{+}$ 的语义，我们直接将轨迹生成与选择条件化于 $z_t^{+}$：
Rather than supervising $z_{t}^{+}$ with pixel-level reconstruction or image-latent prediction objectives, we directly condition trajectory generation and selection on $z_{t}^{+}$:
\begin{equation}
    \tau_{1:T} \sim \pi_{\phi}(h_{t}, z_{t}^{+}).
    \label{eq:posterior_policy}
\end{equation}
%因此，future latent space 由 trajectory generation 与 trajectory selection objectives 共同塑造，使 $z_t^{+}$ 优先编码影响规划的未来信息。我们将该表征称为 planning-shaped future representation。
The future latent space is therefore shaped jointly by the trajectory generation and trajectory selection, so that $z_{t}^{+}$ preferentially encodes future information that affects planning. We refer to this representation as a \emph{planning-shaped future representation}.

%然而，未来观测 $v_{t+1:t+T}$、$c_{t+1:t+T}$ 在部署阶段不可获得，因此 PlanWAM 学习一个仅依赖 history representation 的  prior $P_{\theta}$，并由 Hindsight-to-Foresight Distillation 训练以预测 $z_{t}^{+}$。
However, the future observations $v_{t+1:t+T}$ and $c_{t+1:t+T}$ are unavailable at deployment. PlanWAM therefore learns a prior branch $P_{\theta}$, hereafter referred to as the prior, which depends only on the history representation and is trained by Hindsight-to-Foresight Distillation to predict $z_{t}^{+}$:
\begin{equation}
    \hat{z}_{t}^{+} = P_{\theta}(h_{t}),
    \qquad
    \hat{z}_{t}^{+} \approx z_{t}^{+}.
    \label{eq:_prior}
\end{equation}

%预测的 planning-shaped future representation $\hat{z}_{t}^{+}$ 作为前瞻性 planning context，与 compact history representation $h_{t}$ 融合以指导轨迹生成与选择：
The predicted planning-shaped future representation $\hat{z}_{t}^{+}$ is fused with the compact history representation $h_{t}$ as a foresight planning context to guide trajectory generation and selection:
\begin{equation}
  \tau_{1:T} \sim \pi_{\theta}(h_{t}, \hat{z}_{t}^{+}).
  \label{eq:prior_policy}
\end{equation}

%在推理阶段，PlanWAM 的规划过程可以表示为：
At inference, the planning process of PlanWAM can be expressed as:
\begin{equation}
    p(\tau_{1:T}, \hat{z}_{t}^{+} \mid v_{t-n:t}, c_{t-n:t})
    =
    p_{\theta}(\hat{z}_{t}^{+} \mid h_{t})
    \, p_{\theta}(\tau_{1:T} \mid h_{t}, \hat{z}_{t}^{+}).
    \label{eq:planwam_inference}
\end{equation}

\FloatBarrier

\subsection{Recency-Aware History Representation}
\label{sec:history}
\begin{figure}[h]
  \centering
  \includegraphics[width=0.99\linewidth]{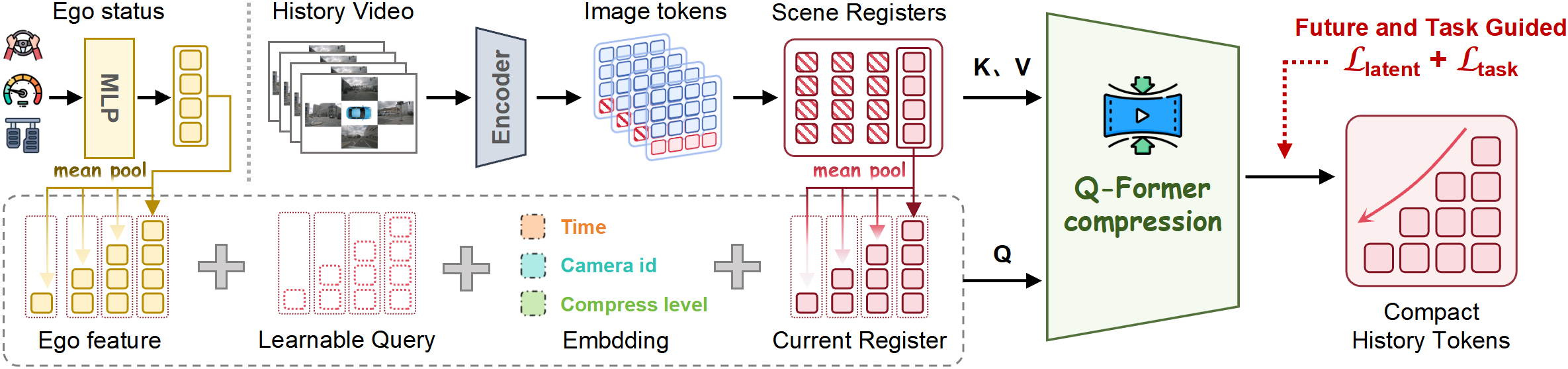}
  %图3：Temporal Register Pyramid for Recency-Aware History Representation.
  \caption{\textbf{Temporal Register Pyramid for Recency-Aware History Representation.}}
  \label{fig:compression}
\end{figure}

%多帧多视角历史观测存在大量时序冗余，且不同时间范围的信息对规划具有不同价值。为此，我们提出 Temporal Register Pyramid，以 recency-aware 的方式逐级压缩历史 scene registers，在有限 token budget 下保留与 future reasoning 和 trajectory planning 相关的动态信息，如图\ref{fig:compression}所示。
Multi-frame, multi-view historical observations contain substantial temporal redundancy, and information from different temporal ranges has different value for planning. To this end, we propose a Temporal Register Pyramid that compresses historical scene registers level by level in a recency-aware manner, retaining, under a limited token budget, the dynamic information relevant to future reasoning and trajectory planning, as illustrated in Fig.~\ref{fig:compression}.

%给定历史观测 $v_{t-n:t}$ 和 ego context $c_{t-n:t}$，共享视觉编码器将时刻 $i$ 的第 $k$ 个相机观测 $v_{i}^{k}$ 编码为 visual tokens，并进一步提取 scene registers $r_{i}^{k}$；自车状态与导航信息 $c_{i}$ 经 MLP 编码为 ego feature $e_{i}$：
Given the historical observations $v_{t-n:t}$ and ego context $c_{t-n:t}$, a shared visual encoder maps the $k$-th camera observation $v_{i}^{k}$ at time $i$ to visual tokens and further extracts scene registers $r_{i}^{k}$; the ego state and navigation information $c_{i}$ are encoded by an MLP into an ego feature $e_{i}$:
\begin{equation}
    x_{i}^{k} = F_{\mathrm{vis}}(v_{i}^{k}),
    \qquad
    r_{i}^{k} = F_{\mathrm{reg}}(x_{i}^{k}),
    \qquad
    e_{i} = F_{\mathrm{ego}}(c_{i}).
    \label{eq:scene_registers}
\end{equation}
%我们根据时间距离将历史划分为 $L$ 个 temporal levels $\{\mathcal{G}_{l}\}_{l=1}^{L}$，并分配不同的 query budget：近期 level 使用更多 learnable queries，较远历史则进行更强压缩。对于第 $l$ 个 level，queries $Q^{(l)}$ 融合 ego feature，以及 time、camera 和 compression-level embeddings；当前时刻的 scene registers 则聚合为共享的 current-scene context $\bar{r}_{t}$：
We partition the history into $L$ temporal levels $\{\mathcal{G}_{l}\}_{l=1}^{L}$ according to temporal distance, and assign different query budgets: recent levels use more learnable queries, whereas more distant history is compressed more strongly. For the $l$-th level, the queries $Q^{(l)}$ are fused with the ego feature as well as time, camera, and compression-level embeddings; the scene registers at the current time are aggregated into a shared current-scene context $\bar{r}_{t}$:
\begin{equation}
    \tilde{Q}^{(l)}
    =
    Q^{(l)}
    + E_{\mathrm{time}}^{(l)}
    + E_{\mathrm{cam}}^{(l)}
    + E_{\mathrm{level}}^{(l)}
    + E_{\mathrm{ego}}^{(l)}
    + \bar{r}_{t}.
    \label{eq:query_augment}
\end{equation}
%随后，我们利用 Q-Former 对各级历史信息进行 query-based compression，以 $\tilde{Q}^{(l)}$ 作为 $Q$，对应 level 的 scene registers $R^{(l)}$ 作为 $K,V$：
We then apply a Q-Former~\citep{li2023blip2} to perform compression of the historical information, using $\tilde{Q}^{(l)}$ as $Q$ and the scene registers $R^{(l)}$ of the corresponding level as $K$ and $V$:
\begin{equation}
    h_{t}^{(l)}
    =
    \mathrm{QFormer}\bigl(Q=\tilde{Q}^{(l)},\, K=R^{(l)},\, V=R^{(l)}\bigr),
    \qquad
    R^{(l)}
    =
    \mathrm{Concat}_{i \in \mathcal{G}_{l},\, k}\, r_{i}^{k}.
    \label{eq:qformer}
\end{equation}
%各 level 的压缩 tokens 最终组成 compact history representation $h_{t}$。我们不采用额外的历史重建目标，而是通过后续 future representation prediction $\mathcal{L}_{\mathrm{latent}}$ 与 trajectory planning objective $\mathcal{L}_{\mathrm{task}}$ 联合优化 $h_{t}$，使其聚焦于未来预测与轨迹规划所需的信息。
The compressed tokens from all levels constitute the compact history representation $h_{t}$. We do not use an additional historical reconstruction objective; instead, $h_{t}$ is jointly optimized by the subsequent future-latent distillation objective $\mathcal{L}_{\mathrm{latent}}$ and the trajectory-planning objective $\mathcal{L}_{\mathrm{task}}$, so that it focuses on the information required for future prediction and trajectory planning.

\subsection{Planning-Shaped Future Representation}
\label{sec:planning}
%给定 Sec.~3.2 的 compact history representation $h_{t}$，我们在训练阶段利用未来观测构造 privileged future posterior。对于未来观测 $v_{t+1:t+T}$，共享视觉编码器以与历史观测相同的方式提取各时刻各相机的 scene registers $r_{i}^{k}$，并加入对应的 future-time embedding $E_{\mathrm{time}}^{\mathrm{fut}}$，形成 future queries：
Given the compact history representation $h_{t}$ from Sec.~\ref{sec:history}, we construct a privileged posterior during training using future observations. For the future observations $v_{t+1:t+T}$, a shared visual encoder extracts scene registers $r_{i}^{k}$ for each time and camera in the same way as for historical observations, and adds the corresponding future-time embedding $E_{\mathrm{time}}^{\mathrm{fut}}$ to form future queries:
\begin{equation}
    f_{t}^{+}
    =
    \mathrm{Concat}_{i,k}\bigl[r_{i}^{k} + E_{\mathrm{time}}^{\mathrm{fut}}(i-t)\bigr],
    \qquad
    i \in \{t+1,\ldots,t+T\}.
    \label{eq:future_queries}
\end{equation}
%不同于使用随机初始化的 learnable queries，privileged posterior $P_{\phi}$ 直接以真实未来 scene registers 构成的 $f_{t}^{+}$ 作为 queries，并以历史表征 $h_{t}$ 作为 context。posterior world model $\mathrm{WM_{\phi}^{\mathrm{post}}}$ 聚合历史证据与真实未来信息，得到 posterior future latent $z_{t}^{+}$：
Rather than using randomly initialized learnable queries, the privileged posterior $P_{\phi}$ takes $f_{t}^{+}$ formed from ground-truth future scene registers directly as queries, and uses the history representation $h_{t}$ as context. The posterior world model $\mathrm{WM_{\phi}^{\mathrm{post}}}$ aggregates historical evidence and ground-truth future information to obtain the posterior future latent $z_{t}^{+}$:
\begin{equation}
    z_{t}^{+}
    =
    P_{\phi}(h_{t}, f_{t}^{+})
    =
    \mathrm{WM_{\phi}^{\mathrm{post}}}\bigl(Q=f_{t}^{+},\, K=h_{t},\, V=h_{t}\bigr).
    \label{eq:zpost}
\end{equation}
%我们不对 future latent $z_{t}^{+}$ 施加额外的通用 reconstruction objective。而是将 $z_{t}^{+}$ 与历史表征 $h_{t}$ 拼接为 posterior planning context $m_{t}^{+}$，并在拼接前加入 type embeddings，以区分 historical evidence 与 future context。得到的 $m_{t}^{+}$ 同时用于 trajectory planner 与 trajectory scorer：
We do not apply an additional generic reconstruction objective to the future latent $z_{t}^{+}$. Instead, we concatenate $z_{t}^{+}$ with the history representation $h_{t}$ to form a posterior planning context $m_{t}^{+}$, adding type embeddings before concatenation to distinguish historical evidence from future context. The resulting $m_{t}^{+}$ is used for both trajectory planner and trajectory scorer:
\begin{equation}
    m_{t}^{+}
    =
    \mathrm{Concat}\bigl(h_{t} + E_{\mathrm{type}}^{\mathrm{hist}},\, z_{t}^{+} + E_{\mathrm{type}}^{\mathrm{fut}}\bigr),
    \quad
    \tau_{1:T}^{+}
    =
    G_{\mathrm{plan}}(m_{t}^{+}),
    \quad
    s^{+}
    =
    G_{\mathrm{score}}(\tau_{1:T}^{+},\, m_{t}^{+}).
    \label{eq:mplan}
\end{equation}
%posterior  由 trajectory generation 与 trajectory selection objectives 监督：
The posterior is supervised by the trajectory generation and trajectory selection objectives:
\begin{equation}
    \mathcal{L}_{\mathrm{post}}
    =
    \mathcal{L}_{\mathrm{plan}}
    +
    \mathcal{L}_{\mathrm{score}}.
    \label{eq:lpost}
\end{equation}
%因此，规划损失能够经由 planner 与 scorer 反向传播并直接塑造 $z_{t}^{+}$。有限的 future latent capacity 因而优先保留影响轨迹生成与选择的未来场景信息。我们将该 $z_{t}^{+}$ 称为 planning-shaped future representation。下一节将其作为 privileged target，训练仅依赖历史的  prior。
The planning losses can therefore back-propagate through the planner and the scorer to shape $z_{t}^{+}$ directly. The limited future latent capacity thus preferentially retains future scene information that affects trajectory generation and selection. We refer to this $z_{t}^{+}$ as a \emph{planning-shaped future representation}. The next section uses it as a privileged target to train a history-only prior.

\subsection{Hindsight-to-Foresight Prediction and Planning}
\label{sec:h2f}
%Sec.~3.3 中的 planning-shaped future representation $z_{t}^{+}$ 依赖未来观测，因此在部署阶段无法获得。为此，我们构建仅以历史表征 $h_{t}$ 为条件的  prior $P_{\theta}$，从历史信息中预测对应的 future latent。不同于 posterior 直接以真实未来 scene registers 作为 queries，prior 使用 learnable future queries $Q^{\mathrm{prior}}$。为在 prior 与 posterior latent tokens 之间建立稳定的逐 token 对应关系，我们按照相同的 future-time、camera 和 register-slot 顺序组织 prior queries，并加入相应的结构化 embeddings：
The planning-shaped future representation $z_{t}^{+}$ in Sec.~\ref{sec:planning} relies on future observations and is therefore unavailable at deployment. We thus construct a prior $P_{\theta}$ conditioned solely on the history representation $h_{t}$ to predict it. Unlike the posterior, which directly uses ground-truth future scene registers as queries, the prior uses learnable future queries $Q^{\mathrm{prior}}$. To establish a consistent token-wise correspondence between the prior and posterior latents, we arrange the prior queries in the same order over future times, camera views, and register slots, and augment them with the corresponding structured embeddings:
\begin{equation}
    Q_{i,k,j}^{\mathrm{prior}}
    =
    Q_{0,i,k,j}^{\mathrm{prior}}
    + E_{\mathrm{time}}^{\mathrm{fut}}(i-t)
    + E_{\mathrm{cam}}(k)
    + E_{\mathrm{slot}}(j),
    \qquad
    i \in \{t+1,\ldots,t+T\}.
    \label{eq:prior_queries}
\end{equation}
%其中 $i$、$k$ 和 $j$ 分别表示未来时刻、相机视角和 scene-register slot。随后，prior world model $\mathrm{WM_{\theta}^{\mathrm{prior}}}$ 仅基于 $h_{t}$ 预测 planning-shaped future representation $\hat{z}_{t}^{+}$：
where $i$, $k$, and $j$ index the future time, camera view, and scene-register slot, respectively. The prior world model $\mathrm{WM_{\theta}^{\mathrm{prior}}}$ then predicts the planning-shaped future representation $\hat{z}_{t}^{+}$ from $h_{t}$ alone:
\begin{equation}
    \hat{z}_{t}^{+}
    =
    P_{\theta}(h_{t})
    =
    \mathrm{WM_{\theta}^{\mathrm{prior}}}\bigl(Q=Q^{\mathrm{prior}},\, K=h_{t},\, V=h_{t}\bigr).
    \label{eq:zprior}
\end{equation}
%为将 privileged posterior 所捕获的未来信息迁移至  prior，我们提出 Hindsight-to-Foresight Distillation，直接对齐 prior prediction $\hat{z}_{t}^{+}$ 与 posterior target $z_{t}^{+}$。通过这一对齐， prior 从历史动态中学习预测 posterior 基于真实未来观测形成的 planning-shaped representation，从而将 privileged hindsight 蒸馏为  foresight：
To transfer the future information captured by the privileged posterior to the prior, we propose Hindsight-to-Foresight Distillation, which directly aligns the prior prediction $\hat{z}_{t}^{+}$ with the posterior target $z_{t}^{+}$. Through this alignment, the prior learns to infer from historical dynamics the planning-shaped representation that the posterior derives from ground-truth future observations, thereby distilling privileged hindsight into  foresight:
\begin{equation}
    \mathcal{L}_{\mathrm{latent}}
    =
    \frac{1}{N_{z} d}
    \bigl\|\hat{z}_{t}^{+} - \mathrm{sg}(z_{t}^{+})\bigr\|_{F}^{2}.
    \label{eq:latent_distillation}
\end{equation}
%其中 $\mathrm{sg}(\cdot)$ 表示 stop-gradient，$\|\cdot\|_{F}$ 表示 Frobenius norm，$N_{z}$ 和 $d$ 分别表示 future latent 的 token 数量和每个 token 的特征维度。
where $\mathrm{sg}(\cdot)$ denotes stop-gradient, $\|\cdot\|_{F}$ denotes the Frobenius norm, and $N_{z}$ and $d$ denote the number of tokens in the future latent and the feature dimension of each token, respectively.
%在加入相应的 type embeddings 后，我们将预测得到的 $\hat{z}_{t}^{+}$ 与 history representation $h_{t}$ 拼接，形成 prior planning context。该 context 同时输入轨迹规划器和打分器，得到如下 Foresight-Conditioned Planning 过程：
After adding the corresponding type embeddings, we concatenate the predicted $\hat{z}_{t}^{+}$ with the history representation $h_{t}$ to form a prior planning context. This context is provided to both the trajectory planner and scorer, yielding the following Foresight-Conditioned Planning formulation:
\begin{equation}
    m_{t}^{-}
    =
    \mathrm{Concat}\bigl(h_{t} + E_{\mathrm{type}}^{\mathrm{hist}},\, \hat{z}_{t}^{+} + E_{\mathrm{type}}^{\mathrm{fut}}\bigr),
    \quad
    \tau_{1:T}
    =
    G_{\mathrm{plan}}(m_{t}^{-}),
    \quad
    s
    =
    G_{\mathrm{score}}(\tau_{1:T},\, m_{t}^{-}).
    \label{eq:foresight_planning}
\end{equation}

\subsection{Training and Inference}
\label{sec:train}
%训练。PlanWAM 采用两阶段训练。第一阶段仅优化 privileged posterior planning-shaped future representation，并通过 trajectory generation、trajectory selection 以及轻量级正则项进行监督：
\textbf{Training.} PlanWAM is trained in two stages. The first stage optimizes only the privileged posterior, which learns the planning-shaped future representation from ground-truth future observations under trajectory generation, trajectory selection, and lightweight regularization objectives:
\begin{equation}
    \mathcal{L}_{\mathrm{task}}
    =
    \mathcal{L}_{\mathrm{plan}}
    +
    \mathcal{L}_{\mathrm{score}}
    +
    \lambda_{\mathrm{reg}}\mathcal{L}_{\mathrm{reg}}.
    \label{eq:task_loss}
\end{equation}
%其中，$\mathcal{L}_{\mathrm{reg}}$ 包含用于缓解 trajectory mode collapse 的 diversity regularization 等。第二阶段冻结 posterior，并使用其 checkpoint 初始化 prior 中结构兼容的参数；随后以较低的学习率 fine-tune  prior，其训练目标为：
where $\mathcal{L}_{\mathrm{reg}}$ includes diversity regularization to mitigate latent anti-collapse. In the second stage, we freeze the posterior and use its checkpoint to initialize structurally compatible parameters in the prior. We then fine-tune the prior with a lower learning rate using the objective:
\begin{equation}
    \mathcal{L}_{\mathrm{train}}
    =
    \mathcal{L}_{\mathrm{task}}
    +
    \lambda_{\mathrm{latent}}\mathcal{L}_{\mathrm{latent}}.
    \label{eq:train_objective}
\end{equation}

%推理。推理阶段移除 privileged posterior，仅保留  prior。历史观测首先编码为 $h_{t}$，随后 prior 预测 planning-shaped future representation $\hat{z}_{t}^{+}$，并将其作为 foresight 与 $h_{t}$ 共同用于 trajectory generation 与 scoring。因此，整个推理过程仅依赖历史可观测信息。
\textbf{Inference.} At inference, we remove the privileged posterior and retain only the prior. Historical observations are first encoded into $h_{t}$, after which the prior predicts the planning-shaped future representation $\hat{z}_{t}^{+}$. This representation serves as foresight and is combined with $h_{t}$ for trajectory planner and scorer. Consequently, the entire inference process relies solely on observable history.

\section{Experiments}
\label{sec:experiments}

\subsection{Experimental Setup}
\label{sec:setup}
\paragraph{Datasets and benchmarks.}
%我们在 NAVSIM-v1、NAVSIM-v2 和 HUGSIM 上评估 PlanWAM，以同时考察规划性能与闭环泛化能力。模型在 NAVSIM \texttt{navtrain}（$103{,}288$ 个场景）上训练，并在 \texttt{navtest}（$12{,}146$ 个场景）上评测。NAVSIM-v1 遵循官方评测协议，并以 PDM Score (PDMS) 作为主要指标；NAVSIM-v2 则采用扩展后的 Extended PDM Score (EPDMS)。此外，我们在高真实感闭环模拟器 HUGSIM 上进行 zero-shot closed-loop evaluation，直接部署训练完成的模型而不进行 HUGSIM-specific fine-tuning，并报告 HUGSIM Driving Score (HD-Score) 和 Route Completion (RC)，以评估模型的闭环驾驶性能与跨环境泛化能力。
We evaluate PlanWAM on NAVSIM-v1~\cite{dauner2024navsim}, NAVSIM-v2~\cite{Cao2025CORL}, and HUGSIM~\cite{zhou2024hugsim} to assess both planning performance and closed-loop generalization. Models are trained on NAVSIM \texttt{navtrain} and evaluated on \texttt{navtest}. NAVSIM-v1 follows the official evaluation protocol and uses PDM Score (PDMS) as the primary metric; NAVSIM-v2 uses the Extended PDM Score (EPDMS). We further perform zero-shot closed-loop evaluation on the high-fidelity closed-loop simulator HUGSIM, deploying the trained model directly without HUGSIM-specific fine-tuning, and report HUGSIM Driving Score (HD-Score) and Route Completion (RC) to assess closed-loop driving performance and cross-environment generalization.

\paragraph{Implementation details.}
%PlanWAM 使用来自前、左、右、后四个相机的 $H{=}4$ 帧历史观测；训练期间，posterior 在未来 $4$ 秒时域内以 $1$ 秒间隔读取多视角帧。视觉骨干采用基础权重冻结、LoRA 秩为 $32$ 的 DINOv2 ViT-S，并从每个相机帧提取 $16$ 个 scene registers。Temporal Register Pyramid 在四个 temporal levels 上为每个相机分配 $[4,4,8,16]$ 个 queries，生成包含 $128$ 个 tokens 的历史表征 $h_t$。四个未来时刻、四个相机及每相机 $16$ 个 registers 共同产生 $N_z{=}256$ 个 posterior future-latent tokens。trajectory planner 在同一时域内生成 $K{=}64$ 条候选轨迹，每条包含 $8$ 个 waypoints；trajectory scorer 对候选轨迹进行打分并完成最终选择。两个训练阶段均使用 AdamW，在 $16$ 块 NVIDIA A800 GPU 上训练，每块 GPU 的 batch size 为 $8$。Stage~1 和 Stage~2 的 learning rate 分别为 $1\times10^{-4}$ 和 $1\times10^{-5}$，latent alignment loss 的权重设为 $\lambda_{\mathrm{latent}}{=}0.05$。更多实现与实验细节见补充材料。
PlanWAM takes $T{=}4$ historical frames from four cameras (front, left, right, and rear), while the training-only posterior observes future multi-view frames at $1$-s intervals over a $4$-s horizon. We use a DINOv2 ViT-S backbone with frozen base weights and rank-$32$ LoRA~\cite{hu2022lora}, extracting $16$ scene registers per camera-frame. The Temporal Register Pyramid assigns $[4,4,8,16]$ queries per camera across four temporal levels, producing a $128$-token history representation $h_t$. Four future times, four cameras, and $16$ registers per camera yield $N_z{=}256$ posterior future-latent tokens. Over the same horizon, the trajectory planner generates $K{=}64$ candidate trajectories of $8$ waypoints, and the trajectory scorer ranks them for final selection. Both stages use AdamW~\cite{loshchilov2019adamw} and are trained on $16$ NVIDIA A800 GPUs with a per-GPU batch size of $8$. The learning rates for Stages~1 and~2 are $1\times10^{-4}$ and $1\times10^{-5}$, respectively, and the latent-alignment weight is set to $\lambda_{\mathrm{latent}}{=}0.05$. 
For more details, please refer to the supplementary materials.

\subsection{Main Results}
\label{sec:main}
\paragraph{Open-loop Planning on NAVSIM.}
%表\ref{tab:navsim_v1}与表\ref{tab:navsim_v2}报告 PlanWAM 在 NAVSIM-v1/v2 \texttt{navtest} 上的规划性能。PlanWAM 在 NAVSIM-v1 上达到 $93.8$ PDMS，优于所有对比的学习方法，比 DrivoR 基线高 $0.7$；与此前最强的 world-model-based planner DriveFuture 相比，增益达到 $3.1$。在更具挑战的 NAVSIM-v2 上，PlanWAM 进一步达到 $90.9$ EPDMS，比此前最强学习结果高 $1.0$，比报告的 Human Agent 基线高 $0.6$。PlanWAM 同时取得最优学习结果 NC $99.0$ 与 TTC $98.5$，表明其在安全相关规划指标上也有稳定提升。这些结果表明，planning-shaped future representations 在两个 NAVSIM 基准上都能为轨迹规划提供有效的前瞻信息。
Tables~\ref{tab:navsim_v2} and~\ref{tab:navsim_v1} report the planning performance on the NAVSIM-v1/v2 \texttt{navtest} benchmarks. PlanWAM achieves 93.8 PDMS on NAVSIM-v1, outperforming all learned methods compared and improving the DrivoR baseline by 0.7 points; compared with the strongest prior world-model-based planner, DriveFuture, the gain reaches 3.1 points. On the more challenging NAVSIM-v2 benchmark, PlanWAM further achieves 90.9 EPDMS, exceeding the previous strongest learned result by 1.0 point and the reported Human Agent baseline by 0.6 points. PlanWAM also obtains the best learned NC of 99.0 and TTC of 98.5, with stable gains on safety-related planning metrics. These results show that planning-shaped future representations provide effective foresight for trajectory planning on both NAVSIM benchmarks.

\begin{table}[ht]
  %表1：NAVSIM-v2 \texttt{navtest} 上的表现。$*$ 表示使用强化学习训练。每列最优学习结果加粗。
  \caption{\textbf{Performance on the NAVSIM-v2 \texttt{navtest} benchmark.} The best learned result in each column is shown in \textbf{bold}.}
  \label{tab:navsim_v2}
  \centering
  \setlength\tabcolsep{1.5mm}
  \renewcommand{\arraystretch}{0.9}
  \scriptsize
  \begin{tabular}{l c c c c c c c c c c c}
    \toprule
    Method & Reference & NC$\uparrow$ & DAC$\uparrow$ & DDC$\uparrow$ & TLC$\uparrow$ & EP$\uparrow$ & TTC$\uparrow$ & LK$\uparrow$ & HC$\uparrow$ & EC$\uparrow$ & EPDMS$\uparrow$ \\
    \midrule
    Human Agent & - & 100.0 & 100.0 & 99.8 & 100.0 & 87.4 & 100.0 & 100.0 & 98.1 & 90.1 & 90.3 \\
    \midrule
    \rowcolor{defrow}
    \multicolumn{12}{c}{\textit{E2E-based planners}} \\
    TransFuser~\cite{chitta2022transfuser} & TPAMI'23 & 96.9 & 89.9 & 97.8 & 99.7 & 87.1 & 95.4 & 92.7 & 98.3 & 87.2 & 76.7 \\
    DiffusionDrive~\cite{liao2025diffusiondrive} & CVPR'25 & 98.2 & 95.9 & 99.4 & 99.8 & 87.5 & 97.3 & 96.8 & 98.3 & 87.7 & 84.5 \\
    MeanFuser~\cite{wang2026meanfuser} & CVPR'26 & 98.3 & 97.2 & 99.6 & 99.8 & 87.6 & 97.4 & 97.3 & 98.3 & \textbf{88.2} & 89.5 \\
    \rowcolor{defrow}
    \multicolumn{12}{c}{\textit{VLM-based planners}} \\
    ReCogDrive~\cite{chen2026recogdrive} & ICLR'26 & 98.3 & 95.2 & 99.5 & 99.8 & 87.1 & 97.5 & 96.6 & 98.3 & 86.5 & 83.6 \\
    SGDrive~\cite{li2026sgdrive} & CVPR'26 & 98.6 & 94.3 & 99.5 & \textbf{99.9} & 86.0 & 97.9 & 96.1 & 98.3 & 85.9 & 86.2 \\
    ExploreVLA~\cite{sheng2026explorevla} & ECCV'26 & 98.8 & 96.2 & 99.6 & 99.8 & 87.1 & 98.2 & \textbf{97.8} & 98.3 & 86.8 & 88.8 \\
    \rowcolor{defrow}
    \multicolumn{12}{c}{\textit{World-model-based planners}} \\
    DriveVLA-W0~\cite{li2026drivevla} & ICLR'26 & 98.5 & 99.1 & 98.0 & 99.7 & 86.4 & 98.1 & 93.2 & 97.9 & 58.9 & 86.1 \\
    DriveLaW~\cite{xia2026drivelaw} & CVPR'26 & 98.7 & 96.9 & 99.6 & 99.8 & 87.5 & 98.3 & 97.6 & \textbf{98.4} & 77.4 & 88.6 \\
    DreamerAD~\cite{yang2026dreamerad} & ECCV'26 & 98.0 & 97.2 & 99.5 & 99.8 & 87.8 & 97.4 & 97.5 & 98.3 & 72.4 & 87.7 \\
    Latent-WAM~\cite{wang2026latentwam} & arXiv'26 & 98.1 & 97.3 & 99.6 & 99.8 & 87.7 & 97.3 & 97.6 & 98.1 & 87.3 & 89.3 \\
    DriveFuture~\cite{hong2026drivefuture} & arXiv'26 & 98.8 & 99.1 & 99.6 & \textbf{99.9} & 86.6 & 98.4 & 96.4 & 98.3 & 74.8 & 89.9 \\
    \rowcolor{catrow} PlanWAM (\textbf{ours}) & - & \textbf{99.0} & \textbf{99.2} & \textbf{99.8} & 99.6 & \textbf{90.8} & \textbf{98.5} & 96.5 & 98.1 & 70.5 & \textbf{90.9} \\
    \bottomrule
  \end{tabular}
\end{table}

\paragraph{Zero-shot Closed-loop Generalization on HUGSIM.}
%表\ref{tab:hugsim}报告在 HUGSIM 上、未经模拟器特定微调的 zero-shot 闭环结果。PlanWAM 的平均 RC 为 $53.7$，HD-Score 为 $38.7$，分别比最强基线高 $7.8$ 与 $9.8$。在 Hard / Extreme 子集上增益仍然明显，RC 为 $51.9$ / $37.2$，HD-Score 为 $35.6$ / $22.0$。这些结果表明，PlanWAM 在不同场景难度上都具有较强的 zero-shot 闭环泛化能力。
Table~\ref{tab:hugsim} reports zero-shot closed-loop results on HUGSIM without simulator-specific fine-tuning. PlanWAM achieves 53.7 average RC and 38.7 HD-Score, outperforming the strongest baselines by 7.8 and 9.8 points, respectively. The gains remain pronounced on the Hard / Extreme subsets, where PlanWAM reaches 51.9 / 37.2 RC and 35.6 / 22.0 HD-Score, respectively. These results demonstrate strong zero-shot closed-loop generalization across varying scenario difficulty.
\begin{table}[ht]
  \centering
  \begin{minipage}[t]{0.54\linewidth}
    \centering
    \scriptsize
    \setlength{\tabcolsep}{0.54pt}
    \renewcommand{\arraystretch}{0.92}
    %表2：NAVSIM-v1 \texttt{navtest}。每列最优学习结果加粗。
    \captionof{table}{\textbf{Performance on the NAVSIM-v1 \texttt{navtest} benchmark.}}
    \label{tab:navsim_v1}
    \vspace{2pt}
    \begin{tabularx}{\linewidth}{>{\raggedright\arraybackslash}X *{5}{>{\tiny}c} c}
      \toprule
      Method & NC$\uparrow$ & DAC$\uparrow$ & TTC$\uparrow$ & C$\uparrow$ & EP$\uparrow$ & PDMS$\uparrow$ \\
      \midrule
      Human Agent & 100 & 100 & 100 & 99.9 & 87.5 & 94.8 \\
      \midrule
      \rowcolor{defrow}
      \multicolumn{7}{c}{\textit{E2E-based planners}} \\
      DiffusionDrive~\cite{liao2025diffusiondrive} & 98.2 & 96.2 & 94.7 & 100 & 82.2 & 88.1 \\
      MeanFuser~\cite{wang2026meanfuser} & 98.6 & 97.0 & 95.0 & 100 & 82.8 & 89.0 \\
      iPad~\cite{guo2025ipad} & 98.6 & 98.3 & 94.9 & 100 & 88.0 & 91.7 \\
      DrivoR~\cite{kirby2026drivor} & 98.9 & 98.3 & 96.2 & 100 & 89.1 & 93.1 \\
      \rowcolor{defrow}
      \multicolumn{7}{c}{\textit{VLM-based planners}} \\
      AutoVLA~\cite{autovala2025} & 98.4 & 95.6 & \textbf{98.0} & 99.9 & 81.9 & 89.1 \\
      ReCogDrive~\cite{chen2026recogdrive} & 97.9 & 97.3 & 94.9 & 100 & 87.3 & 90.8 \\
      ExploreVLA~\cite{sheng2026explorevla} & 98.8 & 98.4 & 96.5 & 99.9 & 83.5 & 90.4 \\
      SGDrive~\cite{li2026sgdrive} & 98.6 & 97.8 & 96.2 & 100 & 85.8 & 91.1 \\
      \rowcolor{defrow}
      \multicolumn{7}{c}{\textit{World-model-based planners}} \\
      WoTE~\cite{li2025end} & 98.5 & 96.8 & 94.9 & 99.9 & 81.9 & 88.3 \\
      Epona~\cite{zhang2025epona} & 97.9 & 95.1 & 93.8 & 99.9 & 80.4 & 86.2 \\
      DreamerAD~\cite{yang2026dreamerad} & 98.0 & 97.2 & 94.3 & 100 & 83.1 & 88.7 \\
      DriveLaW~\cite{xia2026drivelaw} & \textbf{99.0} & 97.1 & 96.7 & 100 & 81.3 & 89.1 \\
      DriveVLA-W0~\cite{li2026drivevla} & 98.7 & \textbf{99.1} & 95.3 & 99.3 & 83.3 & 90.2 \\
      DriveFuture~\cite{hong2026drivefuture} & 98.8 & \textbf{99.1} & 95.4 & 100 & 84.2 & 90.7 \\
      \rowcolor{catrow} PlanWAM (\textbf{ours}) & 98.9 & 98.8 & 96.3 & 100 & \textbf{91.0} & \textbf{93.8} \\
      \bottomrule
    \end{tabularx}
  \end{minipage}
  \hfill
  \begin{minipage}[t]{0.44\linewidth}
    \centering
    \scriptsize
    \setlength{\tabcolsep}{1.6pt}
    \renewcommand{\arraystretch}{0.92}
    %表3：HUGSIM。(a) Route Completion 在上，(b) HD-Score 在下。E/M/H/X 表示 Easy/Medium/Hard/Extreme。
    \captionof{table}{\textbf{Closed-loop comparison on HUGSIM.} E, M, H, and X denote Easy, Medium, Hard, and Extreme.}
    \label{tab:hugsim}
    \vspace{2pt}
    \begin{tabular}{lccccc}
      \toprule
      Method & E & M & H & X & Avg. \\
      \midrule
      UniAD~\cite{hu2023planning} & 58.6 & 41.2 & 40.4 & 26.0 & 40.6 \\
      VAD~\cite{jiang2023vad} & 38.7 & 27.0 & 25.5 & 23.0 & 27.9 \\
      LTF~\cite{chitta2022transfuser} & 68.4 & 40.7 & 36.9 & 25.5 & 41.4 \\
      GTRS-Dense~\cite{gtrs} & 64.2 & \textbf{50.0} & 20.7 & 22.3 & 38.0 \\
      Latent-WAM~\cite{wang2026latentwam} & 84.2 & 42.5 & 30.6 & 35.5 & 45.9 \\
      \midrule
      \rowcolor{catrow} PlanWAM (\textbf{ours}) & \textbf{87.6} & 48.3 & \textbf{51.9} & \textbf{37.2} & \textbf{53.7} \\
      \bottomrule
    \end{tabular}
    \par\vspace{1pt}
    \textbf{(a) Route Completion}
    \par\vspace{6pt}
    \begin{tabular}{lccccc}
      \toprule
      Method & E & M & H & X & Avg. \\
      \midrule
      UniAD~\cite{hu2023planning} & 48.7 & 29.5 & 27.3 & 14.3 & 28.9 \\
      VAD~\cite{jiang2023vad} & 24.3 & 9.9 & 10.4 & 8.2 & 12.3 \\
      LTF~\cite{chitta2022transfuser} & 52.8 & 24.6 & 19.8 & 8.1 & 24.8 \\
      GTRS-Dense~\cite{gtrs} & 55.5 & \textbf{39.0} & 11.7 & 14.3 & 28.6 \\
      Latent-WAM~\cite{wang2026latentwam} & 72.5 & 24.0 & 12.2 & 18.1 & 28.9 \\
      \midrule
      \rowcolor{catrow} PlanWAM (\textbf{ours}) & \textbf{75.3} & 32.9 & \textbf{35.6} & \textbf{22.0} & \textbf{38.7} \\
      \bottomrule
    \end{tabular}
    \par\vspace{1pt}
    \textbf{(b) HD-Score}
  \end{minipage}
 \end{table}
 \FloatBarrier

\begin{figure}[h]
  \centering
  \includegraphics[width=1.0\linewidth]{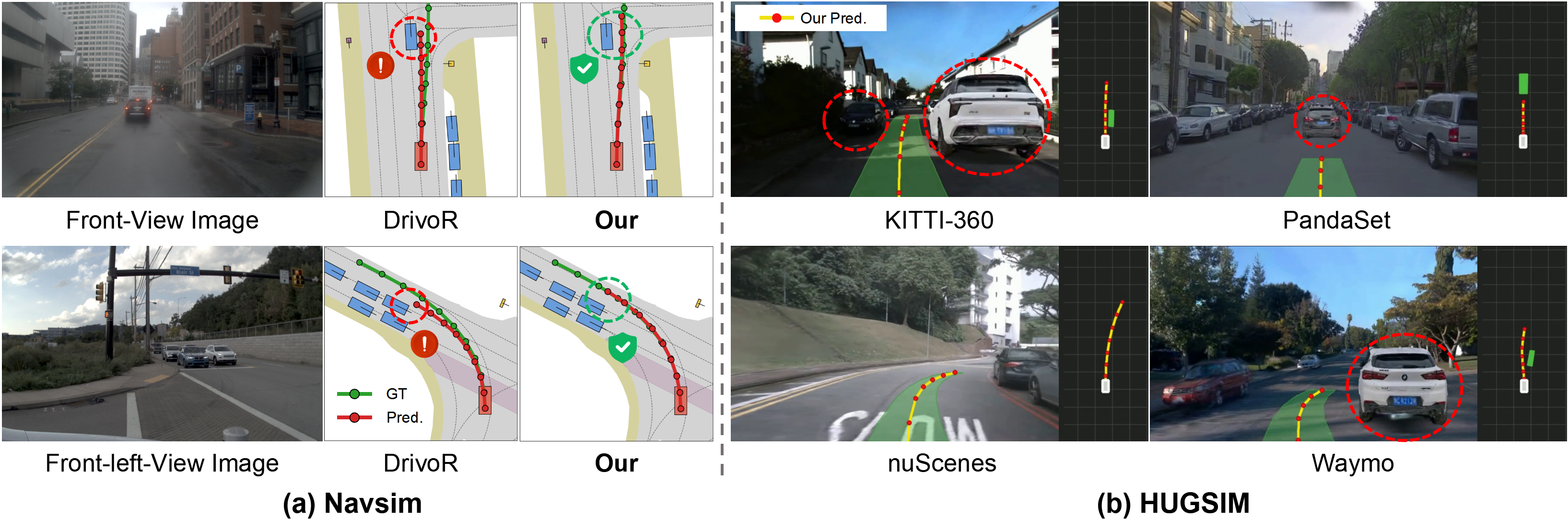}
  %图4：定性结果。(a) NAVSIM 上相对 DrivoR 的对比。(b) HUGSIM 上的 zero-shot 闭环滚动。
  \caption{
  \textbf{Qualitative planning results.}   
  (a) Comparison with DrivoR on representative NAVSIM scenarios. (b) Zero-shot closed-loop rollouts on HUGSIM across different driving domains.}
  \label{fig:qualitative}
\end{figure}

\paragraph{Qualitative Planning Results.}
\label{sec:qual}
%图\ref{fig:qualitative}给出 NAVSIM 与 HUGSIM 上的代表性定性结果。在 NAVSIM 上，相对 DrivoR，PlanWAM 的轨迹更能考虑周围车辆，并在复杂交通场景中避免不安全交互。在 HUGSIM 上，模型零样本迁移到 KITTI-360、PandaSet、nuScenes 与 Waymo，并在多样交通条件下保持稳定的闭环规划。这些例子从定性上支持定量评估中观察到的一致增益。
Fig.~\ref{fig:qualitative} presents representative qualitative results on NAVSIM and HUGSIM. On NAVSIM, compared with DrivoR, PlanWAM produces trajectories that better account for surrounding vehicles and avoid unsafe interactions in challenging traffic scenarios. On HUGSIM, the model transfers zero-shot across KITTI-360, PandaSet, nuScenes, and Waymo, while maintaining stable closed-loop planning in diverse traffic conditions. These examples qualitatively support the consistent gains observed in the quantitative evaluation.

\subsection{Ablation Studies}
\label{sec:ablation}
%我们在 NAVSIM-v1 \texttt{navtest} 上进行系统消融，以分析 PlanWAM 各组成模块的作用。除非特别说明，所有实验均采用完整训练流程，仅改变对应模块。
We conduct systematic ablations on NAVSIM-v1 \texttt{navtest} to analyze the role of each PlanWAM component. Every run follows the full training procedure and changes only the module under study.

\paragraph{Overall Component Evaluation.}
%我们逐步加入 PlanWAM 的三个核心组件：Temporal Register Pyramid (TRP)、Hindsight-to-Foresight Distillation (H2F) 和 Foresight-Conditioned Planning (FCP)，结果见表\ref{tab:ablation_main}。从基础模型开始，引入 TRP 后 PDMS 从 $93.1$ 提升至 $93.4$；进一步加入 H2F 后达到 $93.5$；最终通过 FCP 将预测的 future latent 作为 planning context 参与轨迹生成与选择，达到 $93.8$。
We add the three core components in sequence, the Temporal Register Pyramid (TRP), Hindsight-to-Foresight Distillation (H2F), and Foresight-Conditioned Planning (FCP), as reported in Table~\ref{tab:ablation_main}. From the base model, TRP raises PDMS from $93.1$ to $93.4$, showing that a recency-aware history representation extracts more useful dynamics from multi-frame observations. Adding H2F further reaches $93.5$, so learning a future latent from privileged future information supplies additional foresight. FCP then uses the predicted future latent as planning context for trajectory generation and selection, and the model reaches $93.8$ PDMS.
\begin{table}[h]
  \centering
  %表：NAVSIM-v1 \texttt{navtest} 上三个核心模块的累加消融。
  \caption{\textbf{Overall component ablation of PlanWAM.}}
  \label{tab:ablation_main}
  \setlength\tabcolsep{1.15mm}
  \renewcommand{\arraystretch}{1.05}
  \scriptsize
  \begin{tabularx}{0.95\linewidth}{c *{3}{>{\centering\arraybackslash}p{16mm}} YYYYY>{\columncolor{pdmscol}\centering\arraybackslash}X}
  \toprule
  ID & TRP & H2F & FCP & NC$\uparrow$ & DAC$\uparrow$ & TTC$\uparrow$ & C$\uparrow$ & EP$\uparrow$ & PDMS$\uparrow$ \\
  \midrule
  0 & $\times$ & $\times$ & $\times$ & 98.3 & \textbf{98.9} & 94.6 & \textbf{100} & 90.5 & 93.1 \\
  1 & $\checkmark$ & $\times$ & $\times$ & 98.8 & 98.8 & 95.3 & \textbf{100} & 90.9 & 93.4 \\
  2 & $\checkmark$ & $\checkmark$ & $\times$ & 98.6 & 98.6 & 95.5 & \textbf{100} & \textbf{91.5} & 93.5 \\
  \rowcolor{catrow} 3 & $\checkmark$ & $\checkmark$ & $\checkmark$ & \textbf{98.9} & 98.8 & \textbf{96.3} & \textbf{100} & 91.0 & \textbf{93.8} \\
  \bottomrule
  \end{tabularx}
  \end{table}

\paragraph{Future Representation Learning and Utilization.}
%表\ref{tab:ablation_future}中，仅用通用 future feature supervision 得到 $93.5$ PDMS；由 trajectory planning objectives 塑造 future latent 后达到 $93.8$。表\ref{tab:ablation_fcp}中，仅作辅助监督或只用于 trajectory generation 时收益有限，用于 trajectory selection 后达到 $93.7$，同时条件化 generation 与 selection 达到 $93.8$。
Table~\ref{tab:ablation_future} shows that generic future-feature supervision reaches only $93.5$ PDMS, so a generic prediction target does not by itself yield planning-relevant information. Shaping the future latent directly with the trajectory planning objectives raises performance to $93.8$ PDMS. Table~\ref{tab:ablation_fcp} then varies where the predicted latent is used. Auxiliary supervision or conditioning only trajectory generation brings little gain, whereas using it for trajectory selection reaches $93.7$ PDMS. Conditioning both generation and selection attains $93.8$ PDMS.
\begin{table}[h]
  \centering
  \begin{minipage}[t]{0.48\linewidth}
    \centering
    \scriptsize
    \setlength{\tabcolsep}{1.2mm}
    \renewcommand{\arraystretch}{0.95}
    %表5：NAVSIM-v1 \texttt{navtest} 上的 future representation。
    \captionof{table}{\textbf{Ablation of Future Representation.}}
    \label{tab:ablation_future}
    \vspace{2pt}
    \begin{tabularx}{\linewidth}{l c >{\columncolor{pdmscol}\centering\arraybackslash}X}
    \toprule
    Future representation & Supervision & PDMS$\uparrow$ \\
    \midrule
    Null future & -- & 93.4 \\
    Generic future latent & $\mathcal{L}_{\mathrm{feature}}$ & 93.5 \\
    \rowcolor{catrow} Planning-shaped latent (ours) & $\mathcal{L}_{\mathrm{plan}}$ + $\mathcal{L}_{\mathrm{score}}$ & \textbf{93.8} \\
    \bottomrule
    \end{tabularx}
    \vspace{2pt}
    %$\mathcal{L}_{\mathrm{feature}}$ 表示未来 scene registers 的重建。
    {\raggedleft\scriptsize \textit{$\mathcal{L}_{\mathrm{feature}}$: reconstruction of future scene registers.}\par}
  \end{minipage}
  \hfill
  \begin{minipage}[t]{0.48\linewidth}
    \centering
    \scriptsize
    \setlength{\tabcolsep}{1.4mm}
    \renewcommand{\arraystretch}{0.95}
    %表6：预测 latent 用于轨迹生成、轨迹选择，或两者。
    \captionof{table}{\textbf{Ablation of Future Latent Usage.}}
    \label{tab:ablation_fcp}
    \vspace{2pt}
    \begin{tabularx}{\linewidth}{>{\raggedright\arraybackslash}X c c >{\columncolor{pdmscol}}c}
    \toprule
    Future latent usage & Generation & Selection & PDMS$\uparrow$ \\
    \midrule
    Auxiliary supervision only & $\times$ & $\times$ & 93.5 \\
    Generation only & $\checkmark$ & $\times$ & 93.5 \\
    Selection only & $\times$ & $\checkmark$ & 93.7 \\
    \rowcolor{catrow} Generation + Selection & $\checkmark$ & $\checkmark$ & \textbf{93.8} \\
    \bottomrule
    \end{tabularx}
  \end{minipage}
\end{table}

\paragraph{History Representation Compression and Future Distillation.}
%表\ref{tab:ablation_pyramid}中，同样 $128$ token 下，recency-aware allocation $[4,4,8,16]$ 达到 $93.8$ PDMS，优于均匀分配与相反分配；增至 $256$ token 仅 $93.6$。表\ref{tab:ablation_h2f}中，仅用 prior 时 latent similarity 为 $0.22$，加入蒸馏后升至 $0.92$ 且 PDMS 为 $93.7$；再结合 posterior initialization 达到 $0.93$ 与 $93.8$ PDMS。
Table~\ref{tab:ablation_pyramid} shows that, under the same $128$-token budget, the recency-aware allocation $[4,4,8,16]$ reaches $93.8$ PDMS, above both the uniform allocation and the reversed allocation. Raising the token count to $256$ yields only $93.6$ PDMS, so how the queries are allocated matters more than enlarging the representation. Table~\ref{tab:ablation_h2f} shows that a prior alone has a latent similarity of $0.22$. Adding latent distillation raises it to $0.92$ and PDMS to $93.7$. With posterior initialization, similarity reaches $0.93$ and PDMS $93.8$.
\begin{table}[h]
  \centering
  \begin{minipage}[t]{0.44\linewidth}
    \centering
    \scriptsize
    \setlength{\tabcolsep}{3pt}
    \renewcommand{\arraystretch}{0.95}
    %表7：Temporal Register Pyramid 的 query allocation。
    \captionof{table}{\textbf{Ablation of Temporal Allocation.}}
    \label{tab:ablation_pyramid}
    \vspace{2pt}
    \begin{tabularx}{\linewidth}{c >{\centering\arraybackslash}X c >{\columncolor{pdmscol}\centering\arraybackslash}X}
    \toprule
    Query allocation & Tokens & Latency (ms)$\downarrow$ & PDMS$\uparrow$ \\
    \midrule
    $[16,16,16,16]$ & \textbf{256} & 380 & 93.6 \\
    $[8,8,8,8]$ & 128 & 343 & 93.4 \\
    $[16,8,4,4]$ & 128 & 346 & 92.9 \\
    \rowcolor{catrow} $[4,4,8,16]$ & 128 & 342 & \textbf{93.8} \\
    $[2,4,8,16]$ & 120 & \textbf{335} & 93.6 \\
    \bottomrule
    \end{tabularx}
    \vspace{2pt}
    %推理延迟：单卡 NVIDIA A800，batch size 1，fp32。
    {\raggedleft\scriptsize \textit{Inference latency on one NVIDIA A800, batch size 1, fp32.}\par}
  \end{minipage}
  \hfill
  \begin{minipage}[t]{0.52\linewidth}
    \centering
    \scriptsize
    \setlength{\tabcolsep}{1.2pt}
    \renewcommand{\arraystretch}{0.95}
    %表8：posterior 初始化与 Hindsight-to-Foresight Distillation。
    \captionof{table}{\textbf{Ablation of future latent distillation.}}
    \label{tab:ablation_h2f}
    \vspace{2pt}
    \begin{tabularx}{\linewidth}{l cc c >{\columncolor{pdmscol}\centering\arraybackslash}X}
    \toprule
    Setting & Posterior init. & $\mathcal{L}_{\mathrm{latent}}$ & Latent sim.$\uparrow$ & PDMS$\uparrow$ \\
    \midrule
    Prior-only & $\times$ & $\times$ & 0.22 & 93.4 \\
    + Posterior init. & $\checkmark$ & $\times$ & 0.46 & 93.5 \\
    + H2F distillation & $\times$ & $\checkmark$ & 0.92 & 93.7 \\
    \rowcolor{catrow} Init. + H2F (ours) & $\checkmark$ & $\checkmark$ & \textbf{0.93} & 93.8 \\
    Privileged posterior (oracle) & -- & -- & -- & \textbf{95.0} \\
    \bottomrule
    \end{tabularx}
    \vspace{2pt}
    %Latent sim. 为 $\hat{z}_{t}^{+}$ 与 $z_{t}^{+}$ 的平均余弦相似度。
    {\raggedleft\scriptsize \textit{Latent sim.: mean cosine similarity between $\hat{z}_{t}^{+}$ and $z_{t}^{+}$.}\par}
  \end{minipage}
\end{table}

\subsection{Analysis of Failure Reduction}
\label{sec:failure}
\begin{figure}[h]
   \centering
   \includegraphics[width=0.99\linewidth]{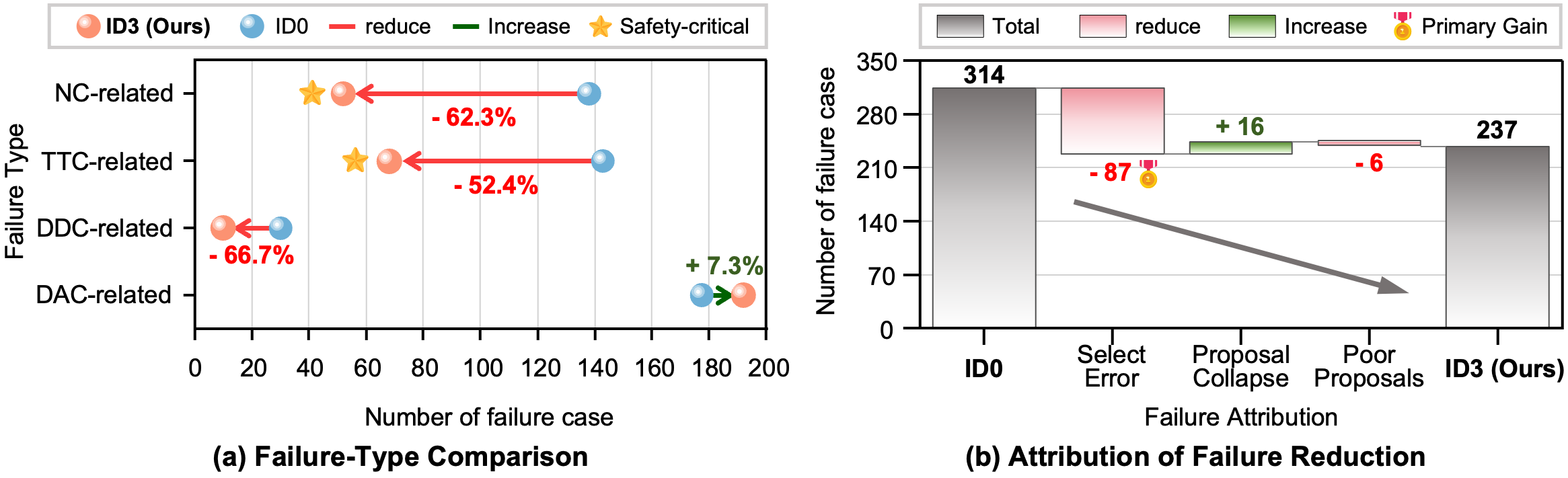}
   %图5：有无世界模型分支的失败分析。(a) 安全关键的 NC、TTC 相关失败显著下降。(b) 下降主要归因于评分误差减少。
   \caption{\textbf{Analysis of failure reduction enabled by planning-shaped foresight.} (a) Failure Type Distribution. (b) Attribution of Failure Reduction.}
   \label{fig:failure_analysis}
\end{figure}

%为进一步理解 planning-oriented foresight 对规划性能提升的作用，我们分析 NAVSIM-v1 \texttt{navtest} 中 PDMS$=0$ 的失败案例。如 Fig.~\ref{fig:failure_analysis}(a) 所示，引入 PlanWAM 后安全相关失败明显减少。其中，No at-fault Collision (NC) 与 Time-to-Collision (TTC) 相关失败分别下降 $62.3\%$ 与 $62.4\%$，表明 planning-shaped future representation 有助于捕获未来交通参与者的运动趋势，从而降低潜在碰撞风险。Driving Direction Compliance (DDC) 相关失败下降 $66.7\%$，进一步说明未来信息改善了动态驾驶约束建模。相比之下，Drivable Area Compliance (DAC) 相关失败基本保持稳定，说明 PlanWAM 的主要收益集中于动态场景理解与决策。如 Fig.~\ref{fig:failure_analysis}(b) 所示，总体失败由 $314$ 降至 $237$，主要来自 trajectory scoring error 减少（$-87$），说明预测的 future latent 提供了有效前瞻，提升了候选轨迹评价与选择的可靠性。proposal quality 也得到一定改善，而 proposal collapse 相关问题仍然存在。该分析表明，PlanWAM 通过学习面向规划的未来表征，为轨迹评价与选择提供有效 foresight，从而减少由未来演化建模不足导致的规划失败。
To further understand how planning-shaped foresight improves planning performance, we analyze the PDMS$=0$ failure cases on NAVSIM-v1 \texttt{navtest}. As shown in Fig.~\ref{fig:failure_analysis}(a), safety-related failures decrease markedly after introducing PlanWAM (ID 0 to ID 3). In particular, No at-fault Collision (NC) and Time-to-Collision (TTC) failures drop by $62.3\%$ and $52.4\%$, indicating that the planning-shaped future representation helps the model capture the motion trends of future traffic participants and thereby lowers potential collision risk. Driving Direction Compliance (DDC) failures also drop by $66.7\%$, further showing that future information improves the modeling of dynamic driving constraints. In contrast, Drivable Area Compliance (DAC) failures show a slight deterioration, indicating that the gains of PlanWAM concentrate on dynamic scene understanding and decision making. As shown in Fig.~\ref{fig:failure_analysis}(b), the number of failures falls from $314$ to $237$, mainly from fewer trajectory select errors ($-87$). The predicted future latent thus provides effective foresight and improves the reliability of candidate-trajectory evaluation and selection. Proposal quality also improves to some extent, while proposal-collapse issues remain. This analysis indicates that, by learning a planning-shaped future representation, PlanWAM provides effective foresight for trajectory evaluation and selection, thereby reducing planning failures caused by insufficient modeling of future evolution.

\section{Conclusion}
%xxx
We presented PlanWAM, a planning-shaped world action model that reformulates future modeling in end-to-end autonomous driving as planning-oriented predictive representation learning. By shaping future latent representations with trajectory planning objectives and distilling them from privileged future observations to a history-only prior, PlanWAM learns deployable foresight for trajectory generation and selection. PlanWAM achieves 93.8 PDMS / 90.9 EPDMS on NAVSIM-v1/v2 navtest and 38.7 HD-Score in zero-shot closed-loop HUGSIM evaluation, demonstrating the effectiveness of planning-shaped future representations for end-to-end autonomous driving.

% ICLR 2027：Reproducibility 强烈建议；AI Use 为强制小节。二者均不计正文页数，放在结论之后、参考文献之前。
\section*{Reproducibility Statement}
%方法、数据划分和主要实现见正文。附录 A 为架构与数据流，B 为训练目标、损失权重和优化设置，C 为 PDMS、EPDMS 与 HUGSIM 的计分协议，D 为补充消融，E 为更多定性结果。实验使用公开的 NAVSIM-v1/v2 与 HUGSIM。匿名仓库含项目概述、可视化视频、补充材料和源代码。
The method of PlanWAM is specified in Sec.~\ref{sec:method}, including the Temporal Register Pyramid, the planning-shaped future posterior, Hindsight-to-Foresight Distillation, and the two-stage training and history-only inference procedure. Dataset splits, evaluation protocols, and the main implementation settings are given in Sec.~\ref{sec:setup}. Appendix~\ref{app:arch} records the architecture and data flow; Appendix~\ref{app:training} the training objectives, loss weights, and optimization schedule; Appendix~\ref{app:benchmarks} the PDMS, EPDMS, and HUGSIM scoring protocols used for the reported numbers; Appendix~\ref{app:ablations} the additional controlled ablations; and Appendix~\ref{app:qualitative} further qualitative results. All experiments use the public NAVSIM-v1, NAVSIM-v2, and HUGSIM benchmarks. An anonymized repository containing a project overview, visualization videos, supplementary material, and source code is available at \url{https://jinchan9.github.io/PlanWAM_project_page}.

\section*{AI Use Statement}
%仅对应投稿表勾选的两项：辅助或润色写作；检索与发现相关工作。未用于起草章节、研究构思或执行、合成数据、数学证明。作者核对全部辅助文本与被引文献，并承担责任。
In this work, we used generative AI tools to aid and polish the writing. This included translation from Chinese notes into English, edits for clarity, grammar, and readability, and formatting of tables and \LaTeX. Additionally, we used generative AI tools for retrieval and discovery of related work: they searched for and suggested potentially relevant literature, and the authors verified each item before deciding which works to cite. We have not used generative AI tools to draft sections of the manuscript, to design or execute the research, to generate synthetic datasets, or to formulate or prove mathematical claims, and these uses are not applicable to this work. We have reviewed all AI-assisted text. We take responsibility for the final content of this work, including text and claims produced with the aid of generative AI.

\bibliography{iclr2027_conference}
\bibliographystyle{iclr2027_conference}

\newpage
\appendix

%附录导读，不单列目录。A：三个 Transformer 的数据流和张量尺寸。B：两阶段优化和全部损失。C：NAVSIM-v1/v2 与 HUGSIM 的指标，以及 HD-Score 和 Avg 的算法。D：四组补充消融，分别是后验寄存器、未来时长、蒸馏权重和可训练模块。E：NAVSIM 候选轨迹和 HUGSIM 闭环滚动的补充可视化。
\textbf{The contents of this appendix are as follows:}
\begin{itemize}\itemsep2pt
  \item Appendix~\ref{app:arch} gives the data flow and tensor shapes of the three transformers, for checking the implementation.
  \item Appendix~\ref{app:training} gives the two-stage optimization and the full set of losses.
  \item Appendix~\ref{app:benchmarks} gives the metric definitions for NAVSIM-v1/v2 and HUGSIM, including how HD-Score and the HUGSIM average are computed.
  \item Appendix~\ref{app:ablations} gives four additional ablations: posterior registers, future duration, the distillation weight, and which modules are trained.
  \item Appendix~\ref{app:qualitative} gives additional visualizations of NAVSIM candidate trajectories and HUGSIM closed-loop rollouts.
\end{itemize}

\section{Detailed Architecture and Data Flow}
\label{app:impl}
\label{app:arch}

Fig.~\ref{fig:app_three_transformers} shows the three transformers used at inference: the world model predicts the future latent from prior queries and the history context; the planner decodes candidate trajectories from that latent; the scorer selects one trajectory. The posterior queries that can replace the prior queries are used only while constructing the training target.

%图已改为 Planner Transformer。世界模型：prior 或 posterior query 作 Q，历史 context 作 K/V，输出 future latent。规划器：历史与 future latent 组成 K/V，planner query 作 Q，MLP 解码候选轨迹。打分器：候选轨迹经 MLP 作 Q，同一记忆作 K/V，输出与 ego status 相加后由六个 MLP 得到子分数并选择轨迹。后验 query 只在训练时使用。
\begin{center}
  \centering
  \includegraphics[width=\linewidth]{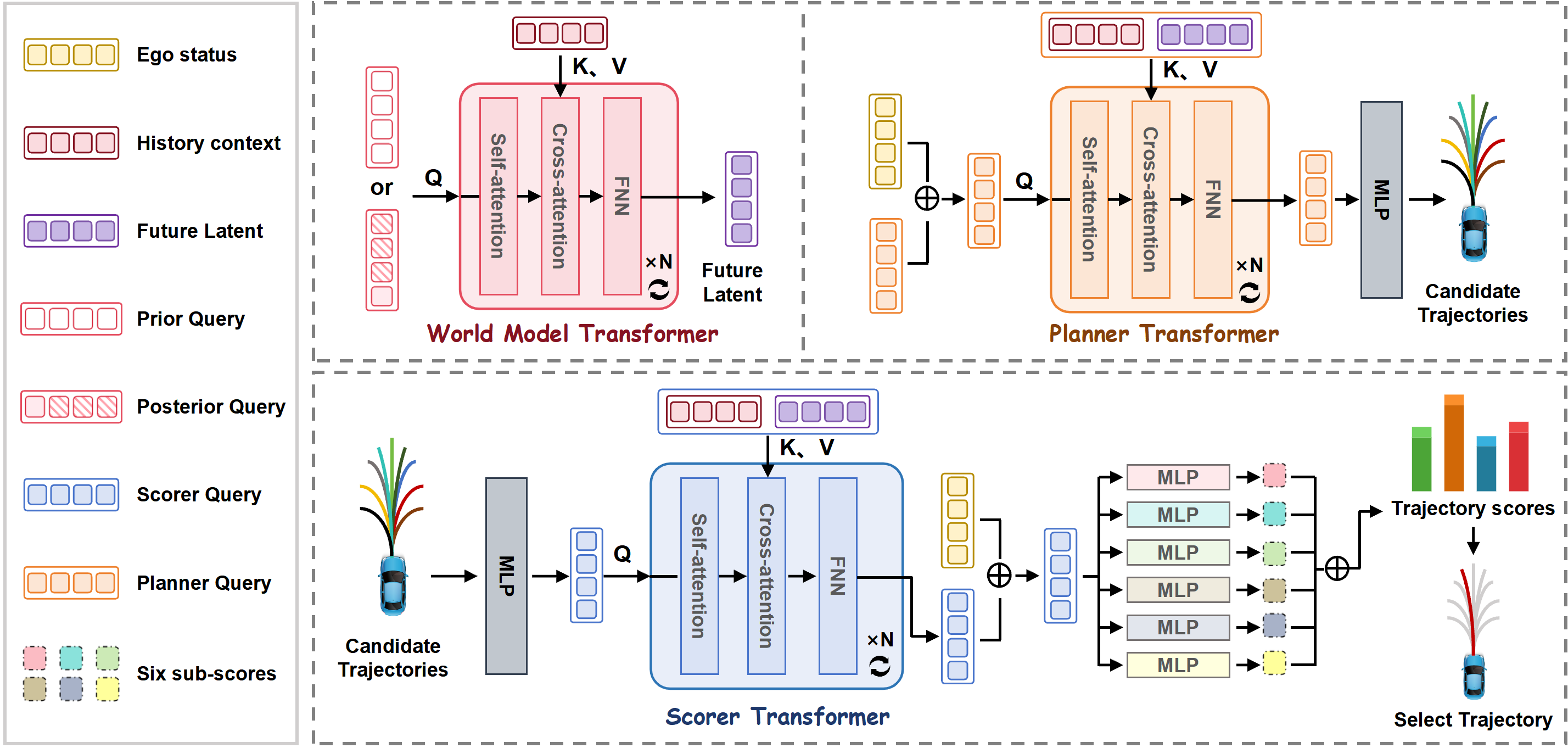}
  \captionof{figure}{\textbf{Detailed transformer architecture and data flow.}
  The world-model transformer takes either posterior queries or prior queries as $Q$, and uses the history context as $K$ and $V$, to produce the future latent. The planner transformer uses planner queries as $Q$ and the combined history--latent memory as $K$ and $V$; an MLP decodes its outputs into candidate trajectories. The scorer transformer embeds those candidates as $Q$, cross-attends to the same memory, adds the ego status, and maps the result through six subscore heads to select one trajectory. Posterior queries are used only during training.}
  \label{fig:app_three_transformers}
\end{center}

\subsection{Token Bookkeeping and End-to-End Flow}
\label{app:token_flow}

The visual backbone extracts $16$ scene registers from each camera frame.  With four historical frames and four cameras, the uncompressed historical stream contains $4\times4\times16=256$ registers.  The Temporal Register Pyramid allocates $[4,4,8,16]$ queries per camera from the oldest to the current temporal level, producing $4\times(4+4+8+16)=128$ history tokens.  The posterior observes four future times, $\{1,2,3,4\}$ seconds, and retains all $4\times4\times16=256$ future slots.  The prior uses the same slot layout, so every predicted token has a fixed time--camera--register counterpart in the posterior target.  Concatenating history and future tokens gives a $384$-token planning memory.

\begin{center}
  \centering
  \captionof{table}{\textbf{Tensor bookkeeping for the reported model.} $B$ is batch size, $d{=}256$, $K{=}64$, and $J{=}8$.}
  \label{tab:app_tensor_shapes}
  \small
  \setlength{\tabcolsep}{4.5pt}
  \begin{tabularx}{\linewidth}{l >{\raggedright\arraybackslash}X l}
    \toprule
    Quantity & Construction & Shape \\
    \midrule
    Raw history registers & $4$ times $\times$ $4$ cameras $\times$ $16$ registers & $B\times256\times d$ \\
    Compressed history $h_t$ & temporal allocation $[4,4,8,16]$ per camera & $B\times128\times d$ \\
    Posterior queries $f_t^+$ & $4$ future times $\times$ $4$ cameras $\times$ $16$ registers & $B\times256\times d$ \\
    Prior queries $Q^{\mathrm{prior}}$ & learned queries with time/camera/slot embeddings & $B\times256\times d$ \\
    Planning memory $m_t^{\pm}$ & history--future concatenation with type embeddings & $B\times384\times d$ \\
    Candidate trajectories & $K$ candidates, $J$ poses $(x,y,\psi)$ & $B\times64\times8\times3$ \\
    Candidate subscores & NC, DAC, TTC, EP, DDC, and comfort & $B\times64\times6$ \\
    \bottomrule
  \end{tabularx}
\end{center}

The data flow is consequently
\begin{equation}
  (v_{t-3:t},c_{t-3:t})
  \xrightarrow{\ E_{\eta}\ } h_t
  \xrightarrow{\ P_{\theta}\ } \hat z_t^+
  \xrightarrow{\ \mathrm{concat}(h_t,\hat z_t^+)\ } m_t^-
  \xrightarrow{\ G_{\mathrm{plan}},G_{\mathrm{score}}\ } \tau^*,
  \label{eq:app_inference_flow}
\end{equation}
whereas Stage~1 replaces $P_{\theta}$ with the privileged posterior $P_{\phi}(h_t,f_t^+)$ and uses the resulting $z_t^+$ to shape the latent space through planning supervision.

\subsection{World-Model Transformer}
\label{app:wm_arch}
%与图一致：Q 为 posterior query 或 prior query，K/V 为历史 context。两者共用四层 pre-norm Transformer，$d{=}256$，4 头，FFN 宽 $1024$。
As in Fig.~\ref{fig:app_three_transformers}, the world-model block is shared in structure by the posterior and the prior. Its $Q$ is either the posterior queries, formed from ground-truth future scene registers, or the learnable prior queries. Its $K$ and $V$ are the history context. Each of the four pre-normalized blocks applies self-attention on the queries, cross-attention from the queries to the history memory, and a feed-forward network, with a residual connection around every sublayer. The hidden width is $d{=}256$, with four attention heads, an FFN width of $1024$, and dropout $0.1$. The block output is the future latent. The two branches differ in the query source:
\begin{align}
  z_t^+ &= \mathrm{WM}_{\phi}^{\mathrm{post}}
  \bigl(Q=f_t^+,K=h_t,V=h_t\bigr), \\
  \hat z_t^+ &= \mathrm{WM}_{\theta}^{\mathrm{prior}}
  \bigl(Q=Q^{\mathrm{prior}},K=h_t,V=h_t\bigr).
\end{align}
The prior queries follow the same future-time, camera, and register-slot order as $f_t^+$, so the token-wise loss in \eqref{eq:latent_distillation} compares corresponding tokens. In the reported model, compatible transformer weights are initialized from the Stage~1 posterior checkpoint; the prior query slots remain prior-specific parameters. At inference only the prior queries are used.

\subsection{Trajectory Planner}
\label{app:planner_arch}
%图中规划器：历史 context 与 future latent 组成 K/V，planner query 作 Q，输出经 MLP 得到候选轨迹。实现上该记忆是 $128+256=384$ 个 token，四层解码，每层 1 个注意力头。
The planner transformer takes planner queries as $Q$. Its $K$ and $V$ are the planning memory formed from the history context and the future latent, the $384$-token concatenation in Table~\ref{tab:app_tensor_shapes}. Four decoder blocks refine the $K{=}64$ candidate tokens. Each block has one-head self-attention among the candidates, one-head cross-attention to the memory, and a width-$1024$ FFN, with hidden width $256$, projection dropout $0.1$, and stochastic depth $0.2$. An MLP ($256\!\rightarrow\!1024\!\rightarrow\!24$) then maps each refined token to $J{=}8$ poses $(x,y,\psi)$, giving the candidate trajectories drawn on the right of Fig.~\ref{fig:app_three_transformers}. The final refinement is passed to the scorer.

\subsection{Trajectory Scorer and Selection}
\label{app:scorer_arch}
%图中打分器：候选轨迹经 MLP 变成 scorer query，作 Q；K/V 仍是历史与 future latent。Transformer 输出与 ego status 相加，六个 MLP 预测子分数，再选出一条轨迹。
The scorer follows the bottom path of Fig.~\ref{fig:app_three_transformers}. Candidate trajectories are embedded by an MLP ($24\!\rightarrow\!1024\!\rightarrow\!256$) and used as $Q$. The $K$ and $V$ are the same history--latent memory used by the planner. Four decoder blocks, again with width $256$, one attention head, and FFN width $1024$, mix the candidate tokens and cross-attend to that memory. The ego-status token is added to every candidate feature. Six MLP heads ($256\!\rightarrow\!1024\!\rightarrow\!1$) then predict NC, DAC, TTC, EP, DDC, and comfort. For NAVSIM-v1 selection, candidate $k$ is ranked by
\begin{equation}
  \log S_k
  = \log p_k^{\mathrm{NC}}+\log p_k^{\mathrm{DAC}}
  +\log\!\left(5p_k^{\mathrm{TTC}}+5p_k^{\mathrm{EP}}+2p_k^{\mathrm{C}}\right),
  \qquad
  k^*=\arg\max_k \log S_k .
  \label{eq:app_candidate_score}
\end{equation}
The constant denominator in PDMS does not change the argmax. DDC is supervised, and it has zero weight in this NAVSIM-v1 ranking. The geometry entering the scorer is detached, so the scoring loss updates the scorer without changing the trajectory decoder through that path.

\subsection{Architecture Summary}
\label{app:architecture_summary}

\begin{center}
  \centering
  \captionof{table}{\textbf{Transformer configurations.} ``Heads'' refers to attention heads in each block; all FFNs use GELU unless the output head is explicitly described as an MLP above.}
  \label{tab:app_architecture}
  \small
  \setlength{\tabcolsep}{3.5pt}
  \begin{tabularx}{\linewidth}{>{\raggedright\arraybackslash}X c c c c >{\raggedright\arraybackslash}X}
    \toprule
    Module & Blocks & Width & Heads & FFN & Output \\
    \midrule
    Temporal Register Pyramid & 2 & 256 & 4 & 1024 & 128 history tokens \\
    Posterior world model & 4 & 256 & 4 & 1024 & 256 target tokens $z_t^+$ \\
    Prior world model & 4 & 256 & 4 & 1024 & 256 predicted tokens $\hat z_t^+$ \\
    Trajectory planner & 4 & 256 & 1 & 1024 & $64\times8\times3$ proposals \\
    Trajectory scorer & 4 & 256 & 1 & 1024 & six scores per proposal \\
    \bottomrule
  \end{tabularx}
\end{center}

\FloatBarrier

\section{Training Objectives and Optimization}
\label{app:training}

\subsection{Two-Stage Training Protocol}
\label{app:optimization}

Table~\ref{tab:app_hparams} gives the complete training schedule used for the reported model.  Stage~1 learns the privileged posterior and the shared planning modules from historical and future observations.  Stage~2 loads structurally compatible parameters from Stage~1, keeps a separate posterior teacher frozen in evaluation mode, and optimizes the history-only prior.  Future images are consumed only by this frozen teacher; they are neither augmented nor exposed to the student.  The teacher output is detached before latent alignment.  At test time, the teacher and all future-observation loading are removed.

% AUTHOR CHECK (manuscript vs launcher): Sec. 4.1 currently reports peak LRs 1e-4/1e-5.
% The inspected launchers default to base_lr=2e-4 (Stage 1) and 1e-4 (Stage 2), with
% sqrt(global_batch/64) scaling. Reconcile these values against the exact reported checkpoints.
\begin{center}
  \centering
  \captionof{table}{\textbf{Optimization and implementation settings.} Learning rates are kept consistent with Sec.~\ref{sec:setup}; all runs use the same camera, token, and trajectory configuration unless an ablation states otherwise.}
  \label{tab:app_hparams}
  \small
  \setlength{\tabcolsep}{4pt}
  \begin{tabularx}{\linewidth}{>{\raggedright\arraybackslash}X >{\raggedright\arraybackslash}X}
    \toprule
    Item & Setting \\
    \midrule
    Input images & four historical frames; front, left, right, rear; $1148\times672$ \\
    Training-only future observations & $\{1,2,3,4\}$ s; same four cameras \\
    Visual encoder & DINOv2 ViT-S; frozen base; LoRA rank $32$ \\
    Planning horizon & $4$ s, $8$ waypoints at $0.5$-s intervals \\
    Candidate / latent counts & $K{=}64$ trajectories; $N_z{=}256$ latent tokens \\
    Optimizer and duration & AdamW; $50$ epochs per stage \\
    Learning-rate schedule & linear warm-up for first $10\%$ of steps, then cosine decay to zero \\
    Peak learning rate & Stage~1: $1\times10^{-4}$; Stage~2: $1\times10^{-5}$ \\
    Hardware and batch & $16\times$ NVIDIA A800; $8$ samples/GPU; global batch $128$ \\
    Regularization weights & $\lambda_{\mathrm{norm}}=\lambda_{\mathrm{div}}=10^{-4}$ \\
    Latent-alignment weight & $\lambda_{\mathrm{latent}}=0.05$ \\
    \bottomrule
  \end{tabularx}
\end{center}

\subsection{Trajectory Generation Objective}
\label{app:plan_loss}

Let $\tau_k\in\mathbb{R}^{J\times3}$ be candidate $k$, $\tau^*$ the demonstrated ego trajectory, and $J{=}8$.  The planner is trained with a best-of-$K$ regression loss,
\begin{equation}
  \mathcal{L}_{\mathrm{plan}}
  =\frac{1}{B}\sum_{b=1}^{B}
  \min_{1\leq k\leq K}
  \frac{1}{J}\sum_{j=1}^{J}
  \left\|\tau_{b,k,j}-\tau^*_{b,j}\right\|_1 .
  \label{eq:app_plan_loss}
\end{equation}
Thus, only the closest candidate receives direct trajectory-regression supervision for a sample, while the learned candidate tokens and self-attention preserve a multimodal proposal set.

\subsection{Trajectory Scoring Objective}
\label{app:score_loss}

Every candidate is evaluated by the official proposal simulator to obtain targets $q_{b,k,m}$ for
\begin{equation*}
  \mathcal{M}=\{\mathrm{NC},\mathrm{DAC},\mathrm{TTC},\mathrm{EP},\mathrm{DDC},\mathrm{C}\}.
\end{equation*}
If $a_{b,k,m}$ denotes the corresponding predicted logit, the scoring loss is
\begin{equation}
  \mathcal{L}_{\mathrm{score}}
  =\sum_{m\in\mathcal{M}}
  \frac{1}{BK}\sum_{b=1}^{B}\sum_{k=1}^{K}
  \operatorname{BCEWithLogits}(a_{b,k,m},q_{b,k,m}).
  \label{eq:app_score_loss}
\end{equation}
Targets taking the intermediate value $0.5$ for NC or DDC are mapped to the conservative negative class for binary training.  Undefined TTC targets are masked rather than treated as negatives.  This loss teaches proposal selection directly from planning-quality signals instead of using distance to the demonstrated trajectory as a proxy score.

\subsection{Latent Regularization and Hindsight-to-Foresight Distillation}
\label{app:loss}

For a latent tensor $z\in\mathbb{R}^{B\times N_z\times d}$, we use two lightweight anti-collapse terms:
\begin{align}
  \mathcal{L}_{\mathrm{norm}}(z)
  &=\frac{1}{BN_z}\sum_{b,i}
  \left[1-\frac{\lVert z_{b,i}\rVert_2}{\sqrt d}\right]_{+}^{2},
  \label{eq:app_norm_reg}\\
  \mathcal{L}_{\mathrm{div}}(z)
  &=\frac{1}{BN_z(N_z-1)}
  \sum_b\sum_{i\neq j}
  \left(
  \frac{z_{b,i}^{\top}z_{b,j}}
  {\lVert z_{b,i}\rVert_2\lVert z_{b,j}\rVert_2}
  \right)^2 .
  \label{eq:app_div_reg}
\end{align}
The first term penalizes vanishing token norms; the second penalizes redundant latent slots.  Written without the shorthand in \eqref{eq:task_loss}, the task objective is
\begin{equation}
  \mathcal{L}_{\mathrm{task}}(z)
  =\mathcal{L}_{\mathrm{plan}}+\mathcal{L}_{\mathrm{score}}
  +10^{-4}\mathcal{L}_{\mathrm{norm}}(z)
  +10^{-4}\mathcal{L}_{\mathrm{div}}(z).
  \label{eq:app_task_expanded}
\end{equation}
Stage~1 substitutes $z=z_t^+$.  Stage~2 substitutes $z=\hat z_t^+$ and adds the token-aligned teacher loss
\begin{equation}
  \mathcal{L}_{\mathrm{Stage\,2}}
  =\mathcal{L}_{\mathrm{task}}(\hat z_t^+)
  +0.05\underbrace{\frac{1}{N_zd}
  \left\|\hat z_t^+-\operatorname{sg}(z_t^+)\right\|_F^2}_{\mathcal{L}_{\mathrm{latent}}}.
  \label{eq:app_stage2_loss}
\end{equation}
The latent loss is computed in FP32 even under mixed-precision training.  The stop-gradient operation and frozen teacher ensure that Stage~2 improves the causal prior rather than moving the privileged target toward the student.

\FloatBarrier

\section{Benchmarks and Evaluation Metrics}
\label{app:benchmarks}
\label{app:metrics}

\subsection{Benchmark Summary}

\begin{center}
  \centering
  \captionof{table}{\textbf{Evaluation benchmarks and their roles.} NAVSIM results use the official \texttt{navtest} evaluator; HUGSIM receives the NAVSIM-trained model without simulator-specific adaptation.}
  \label{tab:app_benchmarks}
  \small
  \setlength{\tabcolsep}{3.5pt}
  \begin{tabularx}{\linewidth}{l >{\raggedright\arraybackslash}X >{\raggedright\arraybackslash}X}
    \toprule
    Benchmark & Data and protocol & Reported metrics \\
    \midrule
    NAVSIM-v1~\cite{dauner2024navsim} & OpenScene/nuPlan-derived real driving logs; non-reactive $4$-s proposal simulation; train on \texttt{navtrain} ($103{,}288$ scenes: $85$k train / $18$k validation), evaluate on \texttt{navtest} ($12{,}146$). & NC, DAC, TTC, EP, C, PDMS \\
    NAVSIM-v2~\cite{Cao2025CORL} & Same split used in this paper, evaluated with the extended official proposal scorer and additional rule/smoothness criteria. & NC, DAC, DDC, TLC, EP, TTC, LK, HC, EC, EPDMS \\
    HUGSIM~\cite{zhou2024hugsim} & Closed-loop rollouts on $436$ public scenarios from nuScenes, KITTI-360, Waymo, and PandaSet; zero-shot transfer of the NAVSIM-trained model. & RC and HD-Score on Easy, Medium, Hard, Extreme, and the scene-weighted average \\
    \bottomrule
  \end{tabularx}
\end{center}

NAVSIM is derived from OpenScene, a compact distribution of nuPlan logs~\cite{caesar2021nuplan}.  It evaluates planned trajectories by rolling the ego vehicle forward while other actors follow recorded trajectories, which makes large-scale evaluation deterministic and non-reactive.  HUGSIM complements this protocol with closed-loop observations: the ego action changes the next rendered view and simulated state, and surrounding actors can interact with the rollout.  The cross-domain HUGSIM test therefore probes compounding planning errors and distribution shift that are not present in a single NAVSIM proposal rollout.

\subsection{NAVSIM-v1: PDM Score}

For each proposal, the official Predictive Driver Model scorer returns no-at-fault collision (NC), drivable-area compliance (DAC), time-to-collision (TTC), normalized ego progress (EP), and comfort (C).  NC and DAC are multiplicative safety gates, whereas TTC, EP, and C form a weighted quality term:
\begin{equation}
  \mathrm{PDMS}
  =\mathrm{NC}\cdot\mathrm{DAC}\cdot
  \frac{5\,\mathrm{TTC}+5\,\mathrm{EP}+2\,\mathrm{C}}{12}.
  \label{eq:app_pdms}
\end{equation}
NC penalizes at-fault collisions along the simulated ego rollout; DAC checks whether the ego footprint stays inside the drivable area; TTC detects imminent projected collisions; EP measures progress along the route centerline and is normalized within the proposal set; and C checks acceleration, jerk, and yaw-related comfort bounds.  Component values and PDMS are first computed per scene and then averaged.  Consequently, the aggregate component means in Table~\ref{tab:navsim_v1} need not reproduce the aggregate PDMS exactly when substituted into \eqref{eq:app_pdms}.

\subsection{NAVSIM-v2: Extended PDM Score}

NAVSIM-v2 adds driving-direction compliance (DDC), traffic-light compliance (TLC), lane keeping (LK), history comfort (HC), and extended comfort (EC).  DDC and TLC join NC and DAC as multiplicative penalties.  LK measures lane-relative stability, HC applies the standard within-trajectory dynamics constraints, and EC measures temporal consistency between consecutive plans.  Before aggregation, the reported evaluator applies a human-penalty filter to every subscore.  Let $\mathcal{M}_{\mathrm{pen}}=\{\mathrm{NC},\mathrm{DAC},\mathrm{DDC},\mathrm{TLC}\}$ and $\mathcal{M}_{\mathrm{avg}}=\{\mathrm{EP},\mathrm{TTC},\mathrm{LK},\mathrm{HC},\mathrm{EC}\}$, with $w_{\mathrm{EP}}=w_{\mathrm{TTC}}=5$ and $w_{\mathrm{LK}}=w_{\mathrm{HC}}=w_{\mathrm{EC}}=2$.  If the human trajectory scores $0$ on a subscore, that term is set to $1$; otherwise the agent's score is used:
\begin{equation}
  \begin{aligned}
    \mathrm{EPDMS}
    &=
    \Biggl(\prod_{m\in\mathcal{M}_{\mathrm{pen}}}
    \mathrm{filter}_{m}\Biggr)
    \cdot
    \frac{\sum_{m\in\mathcal{M}_{\mathrm{avg}}} w_m\,\mathrm{filter}_{m}}
         {\sum_{m\in\mathcal{M}_{\mathrm{avg}}} w_m}, \\
    \mathrm{filter}_{m}
    &=
    \begin{cases}
      1 & \text{if } m(\mathrm{human})=0, \\
      m(\mathrm{agent}) & \text{otherwise.}
    \end{cases}
  \end{aligned}
  \label{eq:app_epdms}
\end{equation}
As for PDMS, the benchmark computes \eqref{eq:app_epdms} at the scene level before averaging; all values in the main tables are displayed on a $0$--$100$ scale.

\subsection{HUGSIM: Route Completion and HD-Score}
%公开评测集共 $436$ 个场景：nuScenes $88$、Waymo $108$、KITTI-360 $113$、PandaSet $127$。难度为 Easy / Medium / Hard / Extreme，分别 $80$ / $157$ / $96$ / $103$ 场。表~\ref{tab:hugsim} 的 Avg. 按这四档场景数加权，等于 $436$ 场的场景平均，不是四档等权，也不是四个数据源等权。
HUGSIM evaluates closed-loop driving on reconstructed scenes from nuScenes, KITTI-360, Waymo, and PandaSet~\cite{zhou2024hugsim}. The public set used here has $436$ scenarios, distributed as in Table~\ref{tab:app_hugsim_split}. Easy inserts no extra vehicle. Medium typically inserts one constant-velocity vehicle. Hard uses more obstructive vehicles, including some attackers. Extreme uses attack planners that intercept the predicted ego future, with more inserted vehicles. We deploy the NAVSIM-trained prior directly, without HUGSIM-specific fine-tuning.

\begin{center}
  \centering
  \captionof{table}{\textbf{HUGSIM public scenario split.} Counts are the $436$ released scenarios.}
  \label{tab:app_hugsim_split}
  \small
  \setlength{\tabcolsep}{5pt}
  \begin{tabular}{lccccc}
    \toprule
    Source & Scenarios & E & M & H & X \\
    \midrule
    nuScenes & 88 & 18 & 34 & 18 & 18 \\
    Waymo & 108 & 22 & 42 & 22 & 22 \\
    KITTI-360 & 113 & 25 & 36 & 26 & 26 \\
    PandaSet & 127 & 15 & 45 & 30 & 37 \\
    \midrule
    Total & 436 & 80 & 157 & 96 & 103 \\
    \bottomrule
  \end{tabular}
\end{center}

Route completion $R_c\in[0,1]$ is the fraction of the prescribed route completed before termination. At each evaluated timestamp $t$, HUGSIM scores no collision, drivable-area compliance, TTC, and comfort. For an episode of $T_e$ timestamps,
\begin{equation}
  \mathrm{HD\text{-}Score}
  =R_c\cdot\frac{1}{T_e}\sum_{t=1}^{T_e}
  \left[
  \mathrm{NC}_t\,\mathrm{DAC}_t
  \frac{5\,\mathrm{TTC}_t+2\,\mathrm{COM}_t}{7}
  \right].
  \label{eq:app_hdscore}
\end{equation}
The product of NC and DAC keeps an off-road or colliding rollout from being offset by comfort, and $R_c$ scales the result by how much of the route is finished. Each difficulty column in Table~\ref{tab:hugsim} is the mean of its scenarios, shown on a $0$--$100$ scale. The Avg.\ column is not $(E+M+H+X)/4$, and it is not an equal average of the four source datasets. It weights the four difficulty means by their scenario counts,
\begin{equation}
  \mathrm{Avg.}
  =
  \frac{80\,E+157\,M+96\,H+103\,X}{436},
  \label{eq:app_hugsim_avg}
\end{equation}
which is the mean over all $436$ scenarios. Medium contains the most scenarios, so it pulls Avg.\ below the unweighted mean of the four difficulties.

\FloatBarrier

\section{Additional Controlled Ablations}
\label{app:ablations}

%四组补充消融均在 NAVSIM-v1 \texttt{navtest} 上报告。D.1 只比较 posterior-only 如何读取未来寄存器。D.2 比较未来时长。D.3 比较蒸馏权重。D.4 比较蒸馏损失以及哪些模块参与更新。
The ablations below are reported on NAVSIM-v1 \texttt{navtest}. This section first compares how the posterior-only model reads future-register tokens, and those scores are posterior PDMS. It then compares the length of the observed future, the weight on latent distillation, and, finally, the distillation loss together with which modules are updated. Each subsection changes the factor named in its title. The highlighted row is the reported setting.

\subsection{Future Registers in the Posterior}
%仅在 posterior-only 分支上比较未来四帧 scene registers 的用法，分数是后验 PDMS，不是部署 prior 的 $93.8$。寄存器直接作 Q 为当前设置，达到 $95.0$；改作 K/V 降到 $94.6$；先用 MLP 压到 $128$ 个 token 再作 Q 降到 $94.2$。
Table~\ref{tab:app_ablation_injection} compares three ways to use the scene-register tokens from the four future frames, on the posterior-only branch. These scores are posterior PDMS, not the deployed prior. Taking the future registers directly as queries is the reported posterior and reaches $95.0$. Using the same tokens as keys and values instead drops the score to $94.6$. Compressing them with an MLP to $128$ tokens and then using those tokens as queries drops it further, to $94.2$. The uncompressed future registers are therefore the stronger queries for the privileged posterior.

\begin{center}
  \centering
  \captionof{table}{\textbf{How the posterior uses future-register tokens.} Scores are posterior-only PDMS on NAVSIM-v1 \texttt{navtest}.}
  \label{tab:app_ablation_injection}
  \small
  \setlength{\tabcolsep}{6pt}
  \begin{tabular}{lcc}
    \toprule
    Future-register role & Tokens & PDMS$\uparrow$ \\
    \midrule
    Keys and values & 256 & 94.6 \\
    \rowcolor{catrow} Queries (ours) & 256 & \textbf{95.0} \\
    MLP compression, then queries & 128 & 94.2 \\
    \bottomrule
  \end{tabular}
\end{center}

\subsection{Observed-Future Duration}
%每个未来时刻是 $4$ 个相机 $\times 16$ 个寄存器，即 $64$ 个 token。只用 $1$ 秒未来为 $64$ token、$93.5$ PDMS；用 $\{1,2,3,4\}$ 秒为 $256$ token、$93.8$ PDMS。更长的未来带来更高分数，说明未来信息对规划有用。
Table~\ref{tab:app_ablation_horizon} varies how many future times the model observes. Each future time contributes one frame from each of the four cameras, and each camera-frame keeps $16$ scene registers, so one future second is $64$ tokens. Using only the $1$\,s future therefore supplies $64$ tokens and reaches $93.5$ PDMS. Extending the observation to $\{1,2,3,4\}$\,s raises the latent to $256$ tokens and the score to $93.8$, which is the reported setting. The gain from the longer horizon shows that the additional future information is useful for planning.
\begin{center}
  \centering
  \captionof{table}{\textbf{Ablation of future duration.} Each future second contributes $4\times 16$ register tokens. Scores are PDMS on NAVSIM-v1 \texttt{navtest}.}
  \label{tab:app_ablation_horizon}
  \small
  \setlength{\tabcolsep}{8pt}
  \begin{tabular}{lcc}
    \toprule
    Future times & Tokens & PDMS$\uparrow$ \\
    \midrule
    $1$\,s & 64 & 93.5 \\
    \rowcolor{catrow} $\{1,2,3,4\}$\,s (ours) & 256 & \textbf{93.8} \\
    \bottomrule
  \end{tabular}
\end{center}

\subsection{Latent-Distillation Weight}
%$\lambda_{\mathrm{latent}}$ 从 $0$ 增到 $0.10$ 时，latent similarity 从 $0.46$ 升到 $0.95$。PDMS 在 $\lambda{=}0.05$ 达到最高 $93.8$，再加大到 $0.10$ 时相似度继续升高，但 PDMS 降到 $93.7$。对齐更紧并不等于规划更好，当前权重取在规划分数的峰值。
Table~\ref{tab:app_ablation_lambda} varies the weight on token-wise latent distillation. With $\lambda_{\mathrm{latent}}{=}0$, the prior is not trained to match the posterior target: latent similarity stays at $0.46$ and PDMS is $93.5$. A small weight of $0.01$ raises similarity to $0.88$ and PDMS to $93.6$. The reported weight, $0.05$, reaches similarity $0.93$ and the highest PDMS, $93.8$. Increasing the weight further to $0.10$ makes the latents still closer, with similarity $0.95$, but PDMS falls back to $93.7$. Tighter alignment therefore does not by itself produce a better plan. The selected weight is the one at which the planning score peaks.

\begin{center}
  \centering
  \captionof{table}{\textbf{Sensitivity to the latent-distillation weight.}}
  \label{tab:app_ablation_lambda}
  \small
  \begin{tabular}{ccc}
    \toprule
    $\lambda_{\mathrm{latent}}$ & Latent sim.$\uparrow$ & PDMS$\uparrow$ \\
    \midrule
    $0$ & 0.46 & 93.5 \\
    $0.01$ & 0.88 & 93.6 \\
    \rowcolor{catrow} $0.05$ (ours) & \textbf{0.93} & \textbf{93.8} \\
    $0.10$ & 0.95 & 93.7 \\
    \bottomrule
  \end{tabular}
\end{center}

\subsection{Distillation Loss and Trainable Modules}
%前两行只训 prior 世界模型：仅 $\mathcal{L}_{\mathrm{latent}}$ 为相似度 $0.96$、PDMS $93.0$；加上 $\mathcal{L}_{\mathrm{plan}}$ 与 $\mathcal{L}_{\mathrm{score}}$ 后为 $0.94$ 与 $93.1$。同一套模块下，相似度再升高并不提高 PDMS：表~\ref{tab:app_ablation_lambda} 中 $\lambda{=}0.05$ 为 $0.93$ / $93.8$，$\lambda{=}0.10$ 为 $0.95$ / $93.7$。后三行在同一损失下扩大可训练模块：冻结 post-only 规划头为 $93.0$--$93.1$，加上 scorer 为 $93.4$，再加 planner 为 $93.5$，除 ViT 基座外更新其余模块为 $93.8$。
Table~\ref{tab:app_ablation_distill_type} varies the distillation loss and the set of trainable modules. The first two rows train only the prior world model. The history encoder, planner, scorer, and ViT stay frozen, and the planner and scorer are copied from the posterior-only model. Under $\mathcal{L}_{\mathrm{latent}}$ alone, latent similarity is $0.96$ and PDMS is $93.0$. Adding $\mathcal{L}_{\mathrm{plan}}$ and $\mathcal{L}_{\mathrm{score}}$, with gradients flowing only into the world model, changes these values to $0.94$ and $93.1$. Latent alignment alone therefore leaves the planning score low. Whether making the latent still closer would raise PDMS is measured with the modules held fixed, in Table~\ref{tab:app_ablation_lambda}: increasing $\lambda_{\mathrm{latent}}$ from $0.05$ to $0.10$ raises similarity from $0.93$ to $0.95$, while PDMS falls from $93.8$ to $93.7$. The planning and scoring losses, not a smaller latent gap, determine the planning score.

The last three rows keep $\mathcal{L}_{\mathrm{latent}}+\mathcal{L}_{\mathrm{plan}}+\mathcal{L}_{\mathrm{score}}$ and train a larger set of modules. With the posterior planner and scorer held fixed, PDMS remains $93.0$--$93.1$. Training the prior world model together with the scorer raises it to $93.4$, and training the planner as well raises it to $93.5$. Training the remaining modules, with the ViT base frozen and adapted by LoRA, reaches $93.8$. Those copied planner and scorer weights were learned for the privileged posterior latent. The prior has to relearn the downstream weights for the latent it predicts.

\begin{center}
  \centering
  \captionof{table}{\textbf{Distillation loss and trainable modules.} The first two rows train only the prior world model and freeze every other module, including LoRA. Scores are PDMS on NAVSIM-v1 \texttt{navtest}.}
  \label{tab:app_ablation_distill_type}
  \small
  \setlength{\tabcolsep}{5pt}
  \begin{tabular}{llcc}
    \toprule
    Trainable modules & Loss & Latent sim.$\uparrow$ & PDMS$\uparrow$ \\
    \midrule
    Prior world model only & $\mathcal{L}_{\mathrm{latent}}$ & \textbf{0.96} & 93.0 \\
    Prior world model only & $\mathcal{L}_{\mathrm{latent}}+\mathcal{L}_{\mathrm{plan}}+\mathcal{L}_{\mathrm{score}}$ & 0.94 & 93.1 \\
    Prior world model and scorer & $\mathcal{L}_{\mathrm{latent}}+\mathcal{L}_{\mathrm{plan}}+\mathcal{L}_{\mathrm{score}}$ & 0.93 & 93.4 \\
    Prior world model, scorer, and planner & $\mathcal{L}_{\mathrm{latent}}+\mathcal{L}_{\mathrm{plan}}+\mathcal{L}_{\mathrm{score}}$ & 0.93 & 93.5 \\
    \rowcolor{catrow} All except the ViT base (ours) & $\mathcal{L}_{\mathrm{latent}}+\mathcal{L}_{\mathrm{plan}}+\mathcal{L}_{\mathrm{score}}$ & 0.93 & \textbf{93.8} \\
    \bottomrule
  \end{tabular}
\end{center}

\FloatBarrier

\section{Additional Qualitative Results}
\label{app:qualitative}

We provide two complementary qualitative views.  The NAVSIM figures expose the complete candidate set and therefore illustrate generation and selection in a single frame.  The HUGSIM figures show consecutive closed-loop observations, revealing whether the selected plan remains stable after its actions alter the subsequent input.  These examples are presented without cherry-picked metric annotations; the quantitative conclusions remain those of Sec.~\ref{sec:main}.

\subsection{NAVSIM Candidate Generation and Selection}

\begin{figure}[ht]
  \centering
  \includegraphics[width=\linewidth]{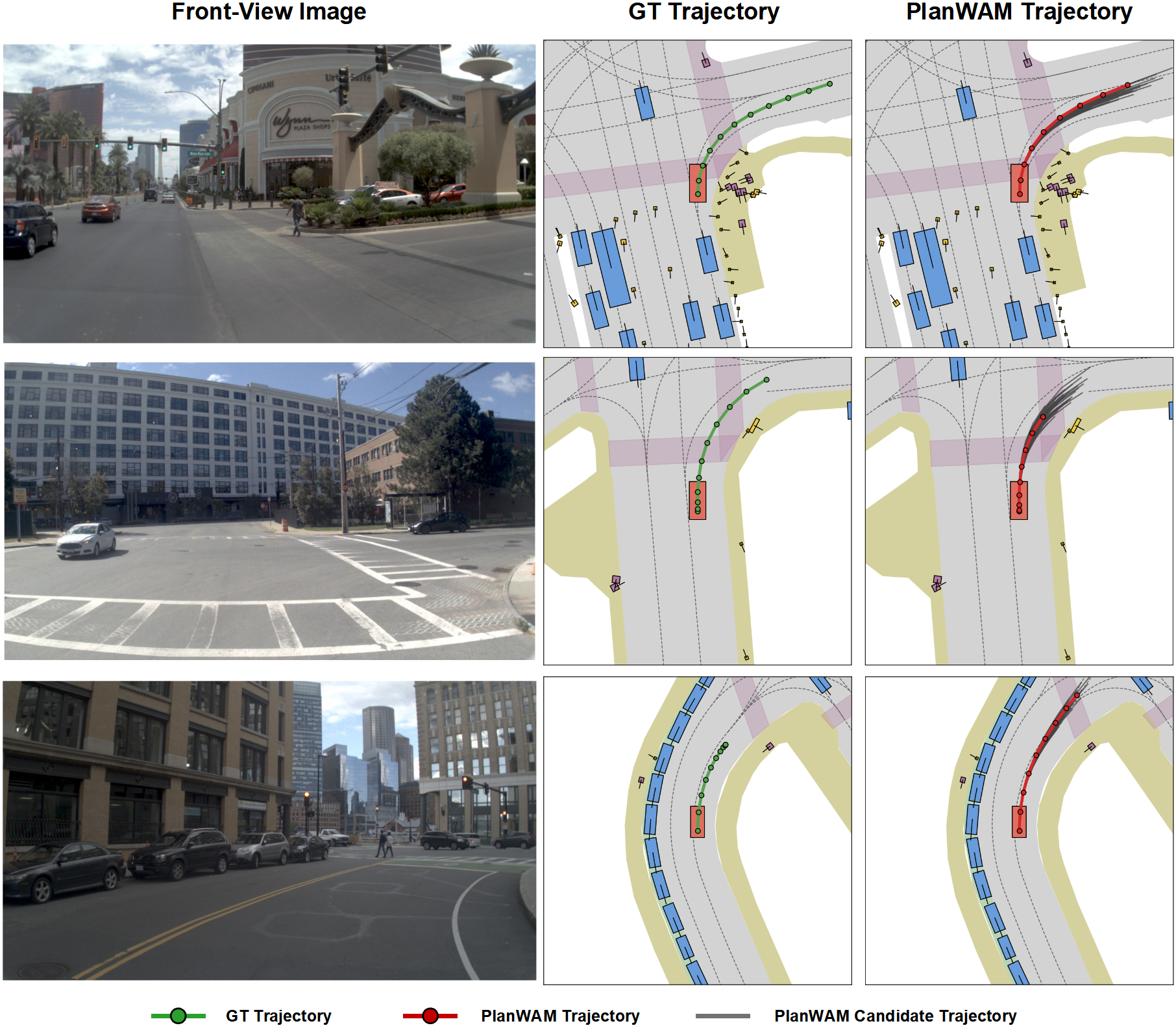}
  \caption{\textbf{Additional NAVSIM results (set 1).} Front view, ground-truth trajectory, and PlanWAM candidates.}
  \label{fig:app_navsim_1}
\end{figure}

\begin{figure}[p]
  \centering
  \includegraphics[width=\linewidth]{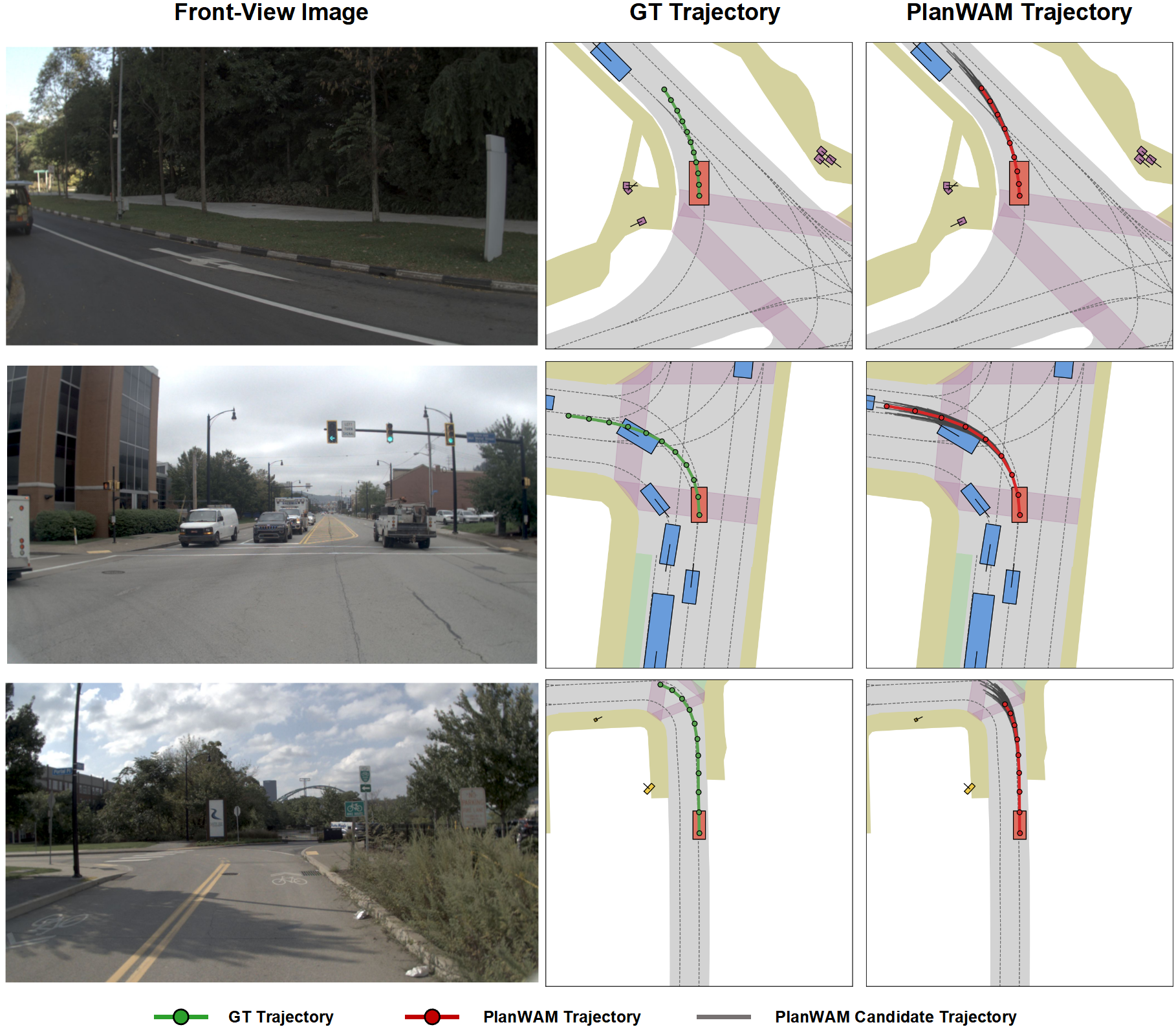}
  \caption{\textbf{Additional NAVSIM results (set 2).} Curved roads and intersections.}
  \label{fig:app_navsim_2}
\end{figure}

\begin{figure}[p]
  \centering
  \includegraphics[width=\linewidth]{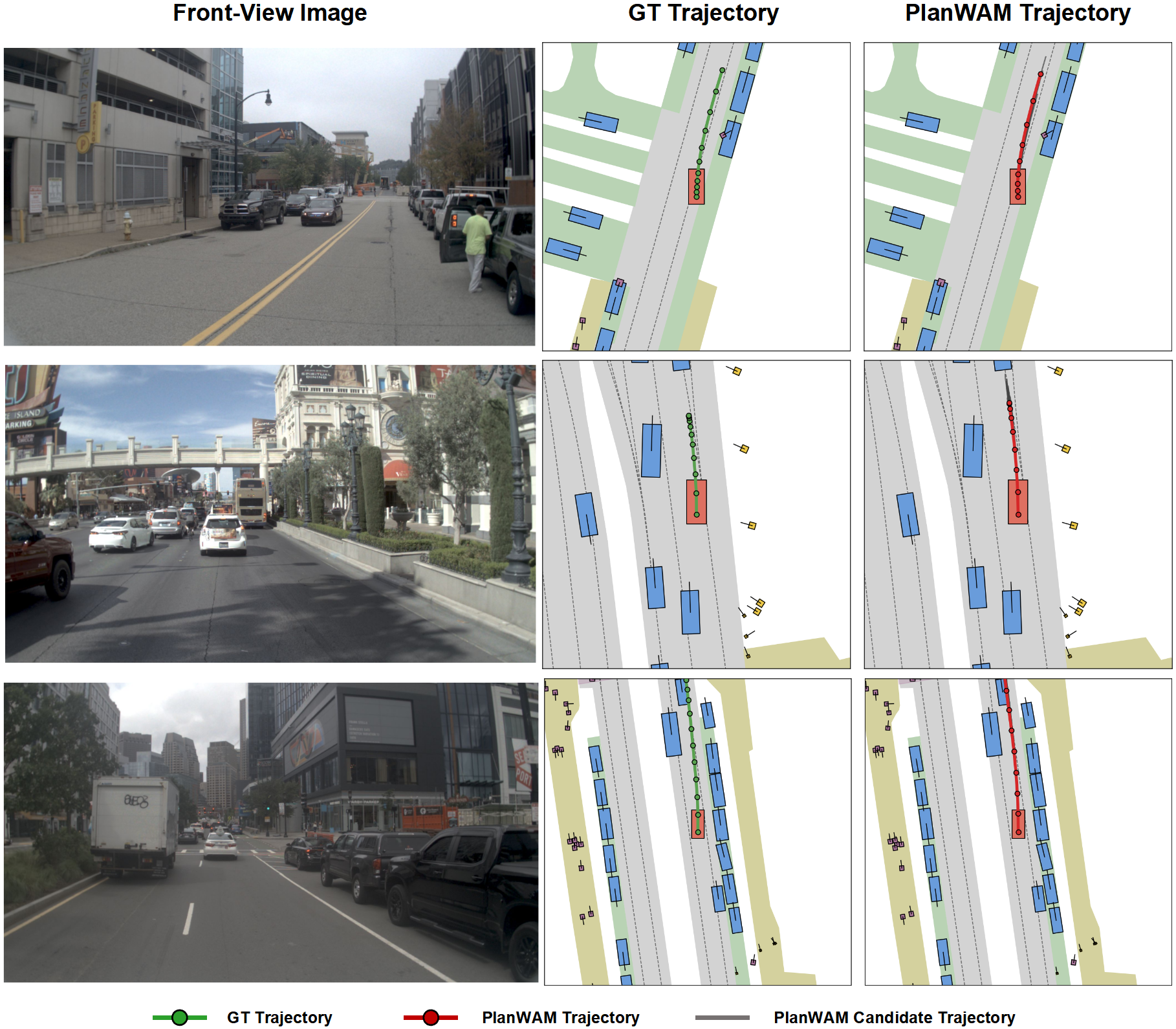}
  \caption{\textbf{Additional NAVSIM results (set 3).} Urban scenes with parked and moving traffic.}
  \label{fig:app_navsim_3}
\end{figure}

\begin{figure}[p]
  \centering
  \includegraphics[width=\linewidth]{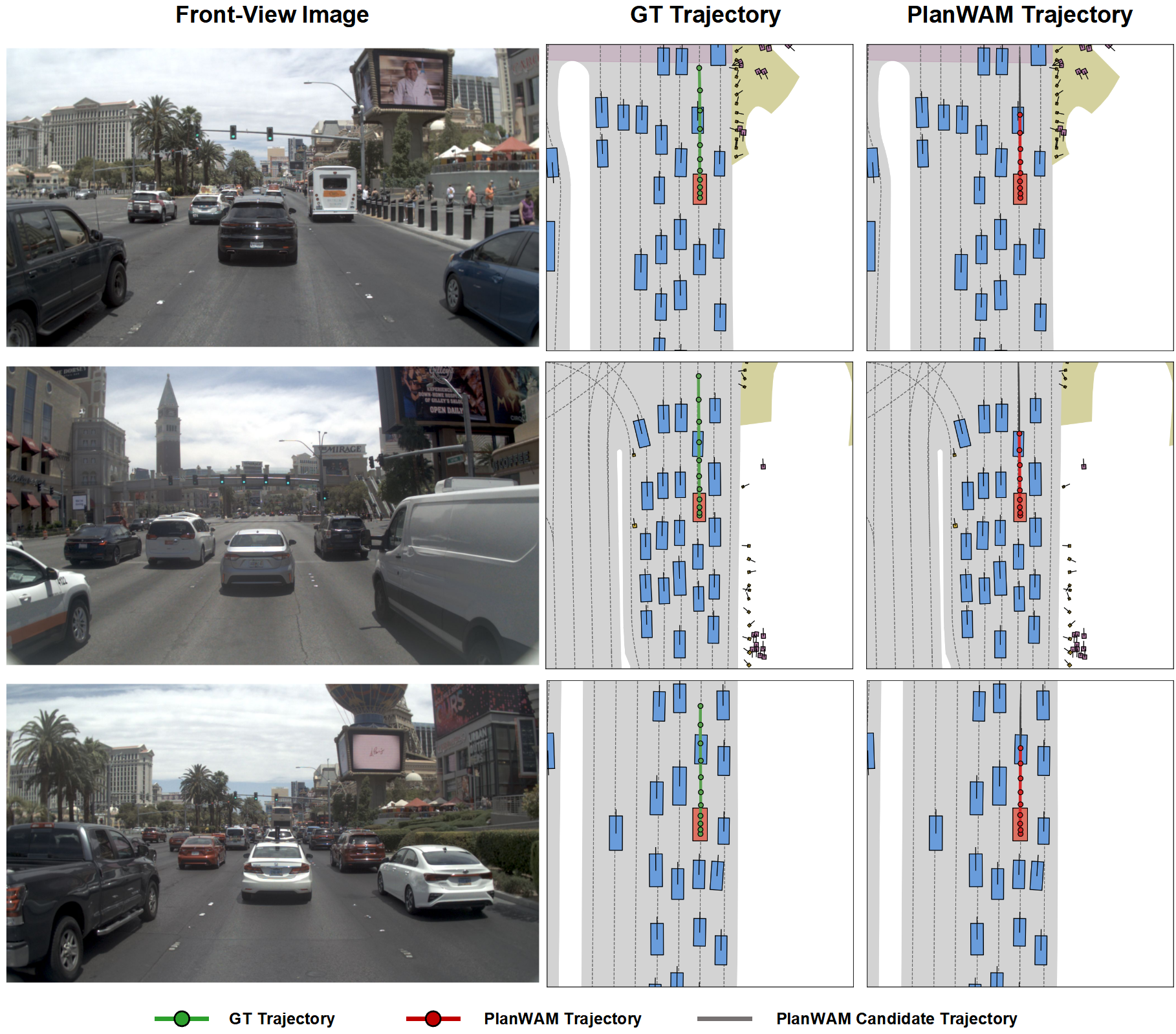}
  \caption{\textbf{Additional NAVSIM results (set 4).} Dense multi-lane traffic. Gray curves are candidates; red is the selected plan.}
  \label{fig:app_navsim_4}
\end{figure}

In Figs.~\ref{fig:app_navsim_1}--\ref{fig:app_navsim_4}, each row contains the front-view image, ground-truth trajectory in the BEV scene, and PlanWAM output.  Gray curves are all $64$ candidates, the red curve is the trajectory selected by PlanWAM, and the green curve is the recorded trajectory.  The cases cover curved roads, turns, intersections, and dense multi-agent traffic.

\clearpage
\subsection{Zero-Shot Closed-Loop Rollouts on HUGSIM}

\begin{figure}[ht]
  \centering
  \includegraphics[width=\linewidth]{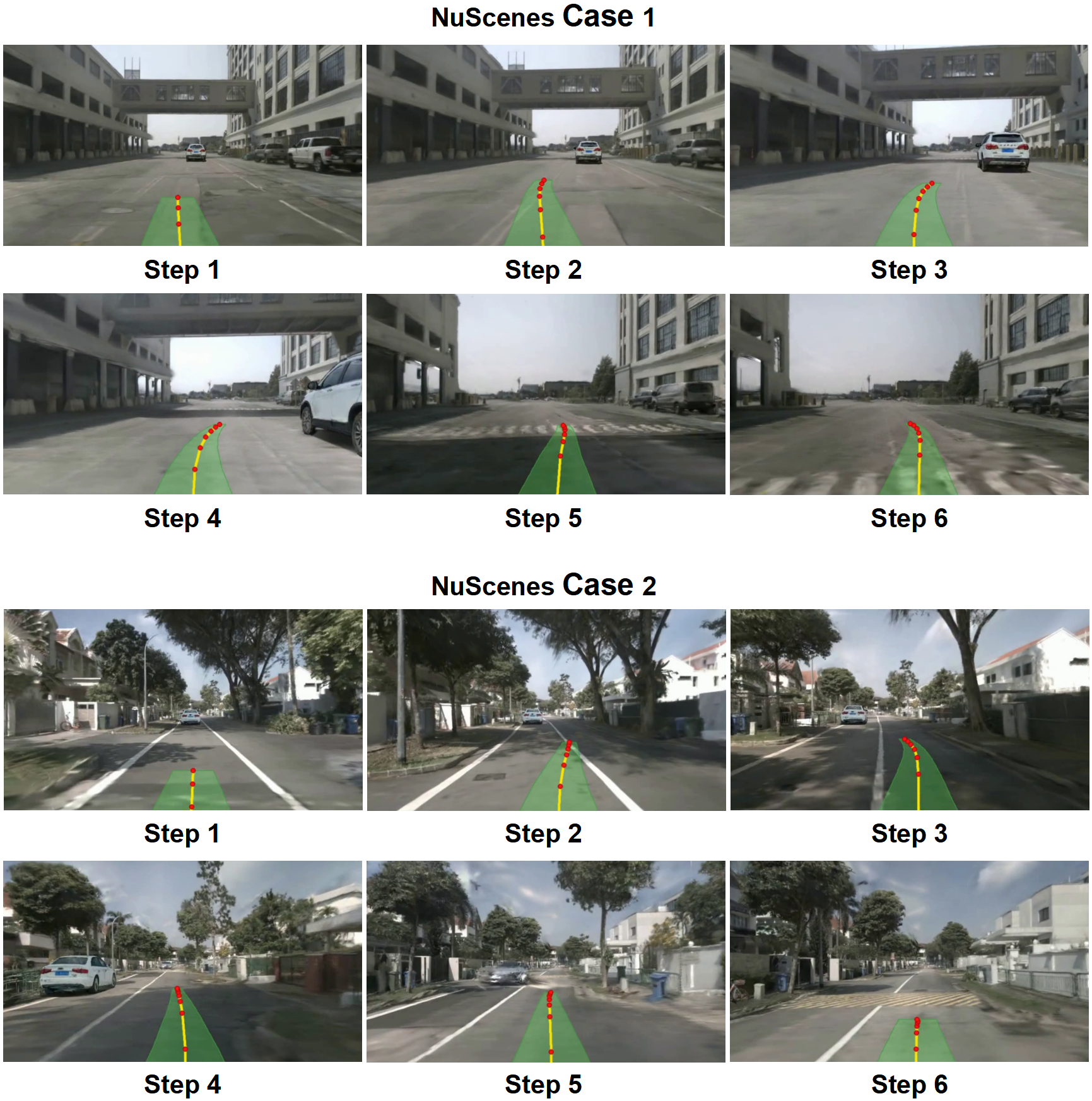}
  \caption{\textbf{Zero-shot HUGSIM rollouts on nuScenes.} Two closed-loop cases are shown over six consecutive planning steps.}
  \label{fig:app_hugsim_nuscenes}
\end{figure}

\begin{figure}[ht]
  \centering
  \includegraphics[width=\linewidth]{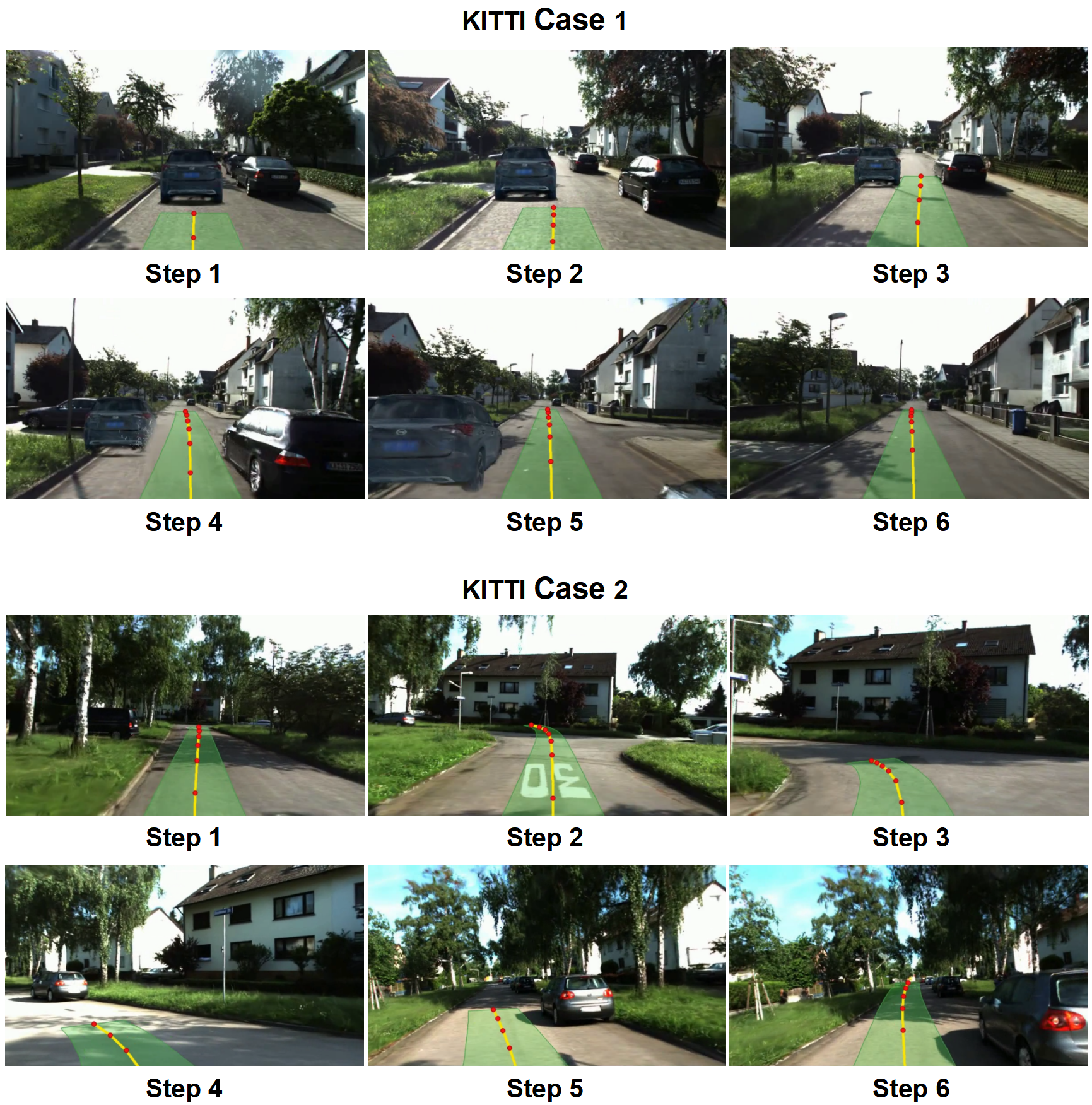}
  \caption{\textbf{Zero-shot HUGSIM rollouts on KITTI-360.} Two closed-loop cases are shown over six consecutive planning steps.}
  \label{fig:app_hugsim_kitti}
\end{figure}

\begin{figure}[ht]
  \centering
  \includegraphics[width=\linewidth]{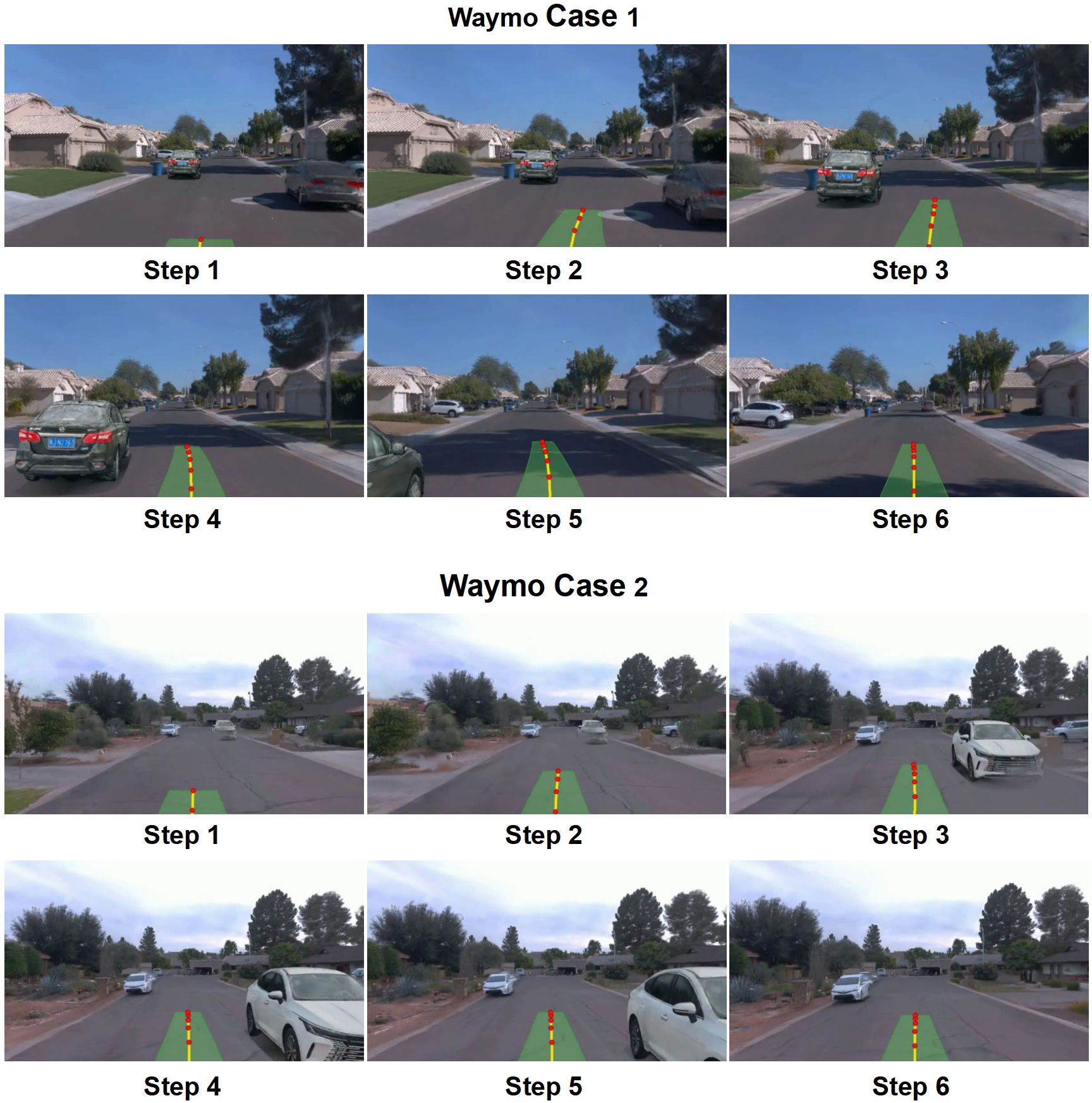}
  \caption{\textbf{Zero-shot HUGSIM rollouts on Waymo.} Two closed-loop cases are shown over six consecutive planning steps.}
  \label{fig:app_hugsim_waymo}
\end{figure}

\begin{figure}[ht]
  \centering
  \includegraphics[width=\linewidth]{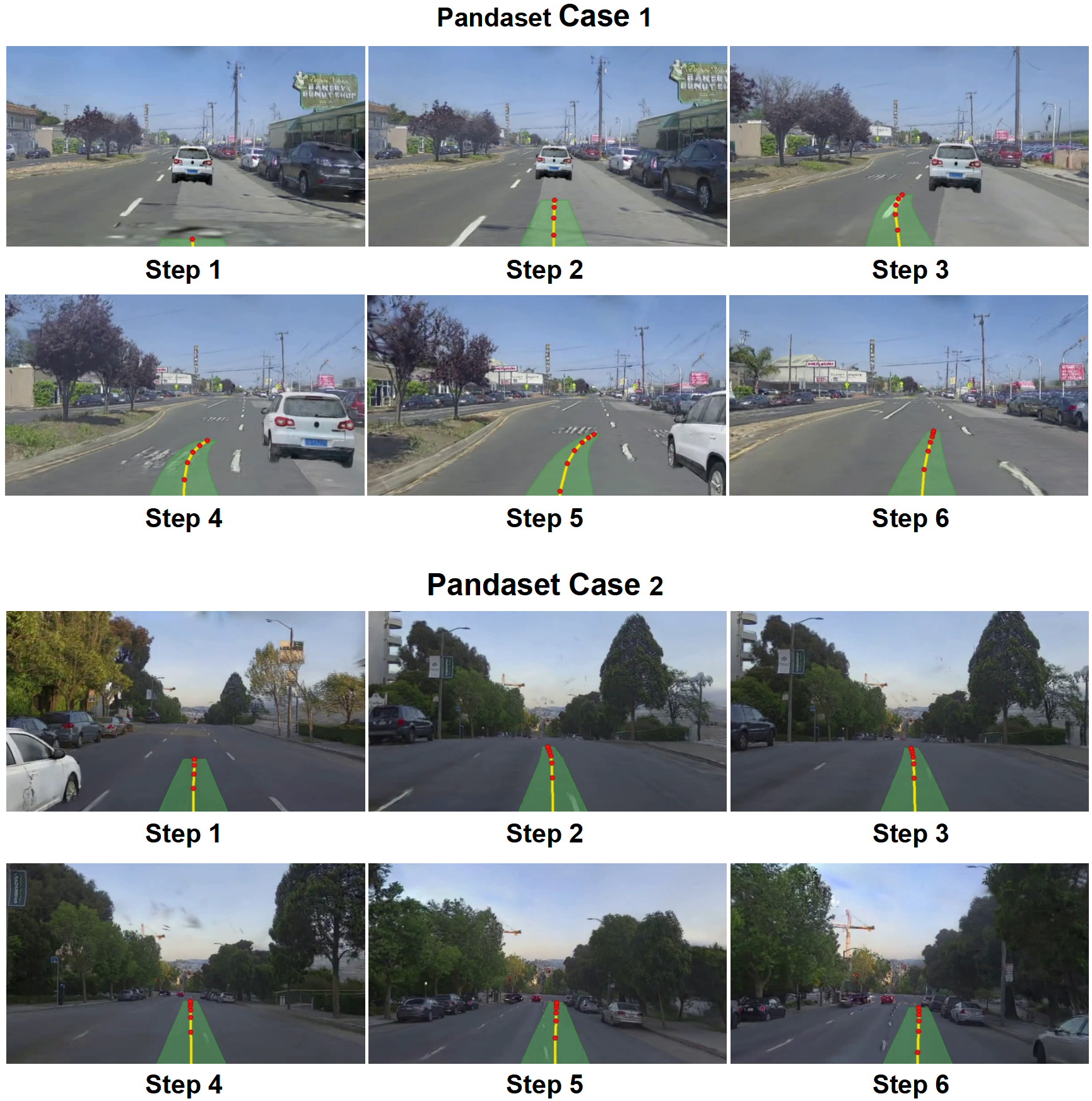}
  \caption{\textbf{Zero-shot HUGSIM rollouts on PandaSet.} Two closed-loop cases are shown over six consecutive planning steps.}
  \label{fig:app_hugsim_pandaset}
\end{figure}

Figures~\ref{fig:app_hugsim_nuscenes}--\ref{fig:app_hugsim_pandaset} show two six-step rollouts for each of the four HUGSIM source domains.  The overlaid corridor, centerline, and waypoints visualize the current PlanWAM plan.  The model is unchanged across domains and receives no HUGSIM-specific training.

\clearpage
\end{document}

%% file: math_commands.tex
\usepackage{amsmath,amsfonts,bm}

\def\eqref#1{equation~\ref{#1}}
\def\1{\bm{1}}

\DeclareMathAlphabet{\mathsfit}{\encodingdefault}{\sfdefault}{m}{sl}
\SetMathAlphabet{\mathsfit}{bold}{\encodingdefault}{\sfdefault}{bx}{n}